\documentclass[pdflatex,sn-mathphys-num]{sn-jnl}

\usepackage{graphicx}%
\usepackage{multirow}%
\usepackage{tabularx} 
\usepackage{amsmath,amssymb,amsfonts}%
\usepackage{amsthm}%
\usepackage{mathrsfs}%
\usepackage[title]{appendix}%
\usepackage{xcolor}%
\usepackage{textcomp}%
\usepackage{natbib}
\usepackage{manyfoot}%
\usepackage{booktabs}%
\usepackage{algorithm}%
\usepackage{algorithmicx}%
\usepackage{algpseudocode}%
\usepackage{listings}%
\usepackage{hyperref}
\usepackage{needspace}
\usepackage{enumitem}
\usepackage[utf8]{inputenc}
\usepackage[T1]{fontenc}
\usepackage{url} 
\usepackage{nicefrac}
\usepackage{microtype}

\begin{document}

\title{CoPESD: A Multi-Level Surgical Motion Dataset for Training Large Vision-Language Models to Co-Pilot Endoscopic Submucosal Dissection}


\author[1]{Guankun Wang}\email{gkwang@link.cuhk.edu.hk}
\equalcont{These authors contributed equally to this work.}
\author[1,2]{Han Xiao}\email{xiaohan@pjlab.org.cn}
\equalcont{These authors contributed equally to this work.}
\author[1]{Huxin Gao}\email{huxingao@cuhk.edu.hk}
\author[1,2]{Renrui Zhang}\email{renruizhang@link.cuhk.edu.hk}
\author[1]{Long Bai}\email{b.long@link.cuhk.edu.hk}
\author[3]{Xiaoxiao Yang}\email{yangxiaoxiao10286@giluhospital.com}
\author[3]{Zhen Li}\email{qilulizhen@sdu.edu.cn}
\author*[1,2]{Hongsheng Li}\email{hsli@ee.cuhk.edu.hk}
\author*[1,4]{Hongliang Ren}\email{hlren@ee.cuhk.edu.hk}

\affil[1]{\orgdiv{Department of Electronic Engineering}, \orgname{The Chinese University of Hong Kong}, \state{Hong Kong}, \country{China}}

\affil[2]{\orgname{Shanghai AI Laboratory}, \city{Shanghai}, \country{China}}

\affil[3]{\orgname{The Qilu Hospital of Shandong University}, \city{Jinan}, \country{China}}

\affil[4]{\orgdiv{Department of Biomedical Engineering}, \orgname{National University of Singapore}, \country{Singapore}}


\abstract{With the advances in surgical robotics, robot-assisted endoscopic submucosal dissection (ESD) enables rapid resection of large lesions, minimizing recurrence rates and improving long-term overall survival. Despite these advantages, ESD is technically challenging and carries high risks of complications, necessitating skilled surgeons and precise instruments. Recent advancements in Large Visual-Language Models (LVLMs) offer promising decision support and predictive planning capabilities for robotic systems, which can augment the accuracy of ESD and reduce procedural risks. However, existing datasets for multi-level fine-grained ESD surgical motion understanding are scarce and lack detailed annotations. In this paper, we design a hierarchical decomposition of ESD motion granularity and introduce a multi-level surgical motion dataset (CoPESD) for training LVLMs as the robotic \textbf{Co}-\textbf{P}ilot of \textbf{E}ndoscopic \textbf{S}ubmucosal \textbf{D}issection. CoPESD includes 17,679 images with 32,699 bounding boxes and 88,395 multi-level motions, from over 35 hours of ESD videos for both robot-assisted and conventional surgeries. CoPESD enables granular analysis of ESD motions, focusing on the complex task of submucosal dissection. Extensive experiments on the LVLMs demonstrate the effectiveness of CoPESD in training LVLMs to predict following surgical robotic motions. As the first multimodal ESD motion dataset, CoPESD supports advanced research in ESD instruction-following and surgical automation. The dataset is available at \href{https://github.com/gkw0010/CoPESD}{https://github.com/gkw0010/CoPESD.}}




\maketitle

\section{Introduction}
Advanced surgical robotics has significantly transformed the landscape of minimally invasive surgery (MIS), offering enhanced precision, reduced recovery times, and improved patient outcomes~\cite{rudiman2021minimally}. For MIS in the gastrointestinal (GI) tract, endoscopic submucosal dissection (ESD) emerges as a pivotal procedure for early-stage GI cancer treatment~\cite{chiu2021colonic, bourke2018endoscopic, ono2021guidelines}. Furthermore, robot-assisted ESD is an emerging technique to facilitate the rapid en-bloc resection of large lesions, which is crucial for minimizing recurrence rates and enhancing long-term survival~\cite{maple2015endoscopic, cui2022robotics}. 
\begin{figure}[!htbp]
    \centering
    \includegraphics[width=5in]{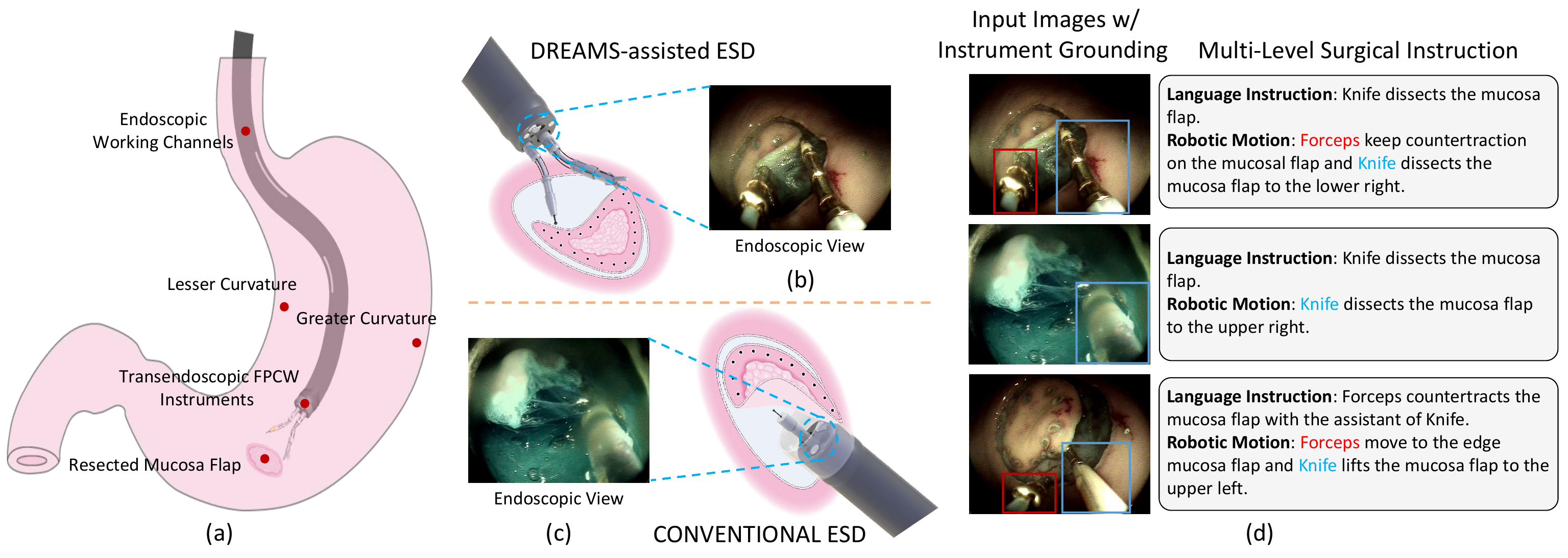}
    \caption{Illustration of Endoscopic Submucosal Dissection with different system instruments. (a) ESD surgery in the gastric body. (b) Endoscopic view of DREAMS-assisted ESD instruments. (c) Endoscopic view of conventional ESD instruments. (d) Multi-level surgical motion instruction demonstrations in CoPESD.} 
    \label{Fig_DREAMS}
\end{figure}
Despite these benefits, the procedure of ESD is still technically challenging, with a high risk of complications such as perforation and bleeding, necessitating a surgeon with exceptional skill and surgical instruments with high dexterity and manipulation accuracy~\cite{odagiri2017complications, yamamoto2009endoscopic}.

In recent years, Large Visual Language Models (LVLMs) have demonstrated superior capabilities in decision support and predictive planning for robotic systems~\cite{li2023vision, wang2024surgical, karamcheti2023language}. Fu et al.~\cite{fu2024multi} have pioneered the LVLMs for surgical task automation. The robot-assisted ESD can also benefit from LVLM's strength based on the availability of a fine-grained vision-language dataset for ESD surgical motions. The LVLM can serve as the co-pilot to augment surgeons' teleoperation accuracy for robot-assisted ESD or endoscope maneuverability (traditional ESD) and further mitigate procedural risks. However, current datasets on robot-assisted or traditional ESD surgical motions are significantly scarce. Although there have been efforts towards the construction of ESD data, existing datasets~\cite{furube2024automated, huang2023experimental, bai2024ossar} are not publicly accessible. The open-source dataset~\cite{cao2023intelligent} only focuses on surgical workflow recognition and lacks detailed annotations for subdivided multi-level fine-grained surgical motions.

To satisfy the demands for a comprehensive vision-language dataset for ESD surgical action instruction-following and motion understanding, we conduct a hierarchical decomposition of ESD surgical motion, which is a critical task for capturing accurate surgical motions. The granularity levels, from high to low, encompass operation, task, surgeme, motion primitive, and navigating motion primitive. Most studies on surgical automation primarily target the task level~\cite{cao2023intelligent}. In contrast, our focus is on the fine-grained motion level in the task of submucosal dissection due to its high degree of soft tissue interaction and complexity in ESD procedures. Consequently, we propose \textbf{CoPESD}, a multi-level surgical motion dataset for training LVLMs as the \textbf{Co}-\textbf{P}ilot of \textbf{ESD}. CoPESD comprises 17,679 images, each annotated with multi-level robotic motions and corresponding bounding boxes, resulting in a total of 32,699 bounding boxes. These images are extracted from over 35 hours of 40 ESD videos, collected using in-vivo porcine models with both robot-assisted~\cite{gao2024transendoscopic} and conventional ESD techniques. Figure~\ref{Fig_DREAMS} illustrates both techniques and procedures performed in gastric stomaches. CoPESD, as the first multimodal ESD surgical motion dataset, serves as a valuable resource for advancing research in ESD motion instruction-following and surgical automation.

Using the proposed CoPESD dataset, we effectively adapt state-of-the-art LVLMs to function as ESD co-pilots to follow surgical instructions. Specifically, images are provided as inputs along with surgeme serving as language instruction prompts. The LVLMs are guided to perceive surgical scenes based on image features and motion text embeddings, thereby outputting low-level robotic motions that align more closely with automated execution. Examples of these outputs are depicted in Figure~\ref{Fig_DREAMS}(d). The low-level motions correspond to the navigating motion primitive within CoPESD. We perform a comprehensive evaluation of the CoPESD dataset utilizing state-of-the-art LVLMs. Quantitative analyses are reported encompassing overall and keyword-specific motion prediction performance, demonstrating the substantial capabilities that CoPESD imparts to LVLMs in the context of surgical motion prediction.

In summary, our contributions are threefold: First, we achieve a granular decomposition of surgical motions, providing precise motion definitions for ESD. Second, we develop CoPESD, a fine-grained multi-level surgical motion dataset, which significantly enhances the resources available for ESD motion instruction-following and surgical automation. Lastly, our comprehensive evaluations of LVLMs demonstrate the significant effectiveness of CoPESD in predicting surgical motions, which allows the LVLM to excel as an ESD co-pilot. As the first multimodal ESD surgical motion dataset, CoPESD is poised to advance research in ESD automation and further integrate AI into minimally invasive surgery.

\section{Related Work}
\subsection{Language Motion Processing in Robotics}
Advancements in natural language processing (NLP) have garnered significant interest in the field of robotics~\cite{tellex2020robots}, particularly in the context of learning groundings between visual and language modalities~\cite{kazemzadeh2014referitgame, lu2019vilbert}. Recent achievements in human-robot interaction encompass the development of an interactive fetching system capable of localizing objects referenced in natural language expressions~\cite{paul2016efficient, shridhar2018interactive, hatori2018interactively, nguyen2020robot, zhang2021invigorate}. Furthermore, these advancements extend to grounding not only objects but also spatial relations, thereby enabling robots to comprehend and execute language-based motion commands~\cite{mees2021composing, venkatesh2021spatial, liu2022structformer}. Prior research on integrating language and vision for motion mapping has predominantly been conducted in constrained environments~\cite{yu2018interactive, misra2017mapping} and has utilized simplified actuators with discrete motion primitives~\cite{anderson2018vision, shridhar2020alfred, shridhar2022cliport}. Recently, there has been an increasing focus on developing language-conditioned policies for continuous visuomotor control in three-dimensional environments, employing methods such as imitation learning~\cite{lynch2020language, stepputtis2020language, jang2022bc} and reinforcement learning~\cite{nair2022learning, shao2021concept2robot, blukis2018mapping}. To establish standardized benchmarks and algorithm implementations, \cite{mees2022calvin} proposes CALVIN. This benchmark incorporates more subtasks and extended long-horizon evaluation sequences, which facilitates the assessment of zero-shot generalization by utilizing unseen language instructions and multi-manipulation environments.

\subsection{Language Motion Processing in Surgical Domain}
The concept of surgical language motion processing originated from the need to create more intuitive and efficient ways for surgeons to control robotic systems during complex procedures. This approach allows for the translation of verbal commands into precise robotic movements, streamlining the surgical work. Pioneering work~\cite{nagy2019dvrk} introduces a DVRK-based framework that integrates language motion processing with the Robot Operating System (ROS) to automate surgical subtasks. Their system leverages stereo vision and hierarchical motion planning to execute commands like knot-tying and blunt dissection, effectively reducing the cognitive load on surgeons. Ginesi et al.~\cite{ginesi2021dynamic} further contribute to the field of robotic surgery by exploring situation awareness and autonomous task planning. They highlighted the role of dynamic motion primitive and volumetric obstacle avoidance, enhancing the robot's capability to interpret and execute complex commands. Nguyen et al.~\cite{nguyen2019new} demonstrates the potential of deep reinforcement learning in surgical pattern cutting, emphasizing how AI can be used to refine the execution of natural language instructions. In the field of Endoscopic Submucosal Dissection, various datasets~\cite{furube2024automated, huang2023experimental, bai2024ossar, cao2023intelligent, wang2022real} have been developed to aid in the analysis and automation of this procedure. However, they focus on tasks such as surgical workflow recognition and landmark detection, lack detailed motion granularity, and are mostly not publicly accessible for reasons such as patient privacy. Therefore, we propose CoPESD dataset to address these limitations by providing a comprehensive multimodal dataset with detailed annotations across multi-level surgical motion granularity, from high-level surgical operation captions to low-level navigating motion primitives.

\section{Preliminary}
\subsection{Endoscopic Submucosal Dissection}

ESD is a minimally invasive procedure facilitating the excision of precancerous or early-stage neoplastic lesions within the GI tract, obviating the necessity for open surgical intervention. ESD employs a flexible endoscope to pass through a natural orifice (mouth or anus) and GI tract. The endoscope tip is integrated with a camera to visualize the GI tract for endoscopists. Behind the endoscope tip, the endoscope features a bending section with two degrees of freedom (DoF), pitch and yaw, which are teleoperated by an endoscopist via the proximal handwheels or master console to execute the real-time lesion location~\cite{maple2015endoscopic}. Inside the flexible endoscope, there are one or two working channels for long and flexible instruments (e.g., electrical knife, forceps, injection needle). The traditional ESD instruments have limited dexterity, so their positioning process to targeted lesions generally depends on the maneuverability of the flexible endoscope. With the assistance of the flexible endoscope and instruments, ESD is performed sequentially as the following steps: marking, injection, circumferentially incision and submucosal dissection. During submucosal dissection, lesion countertraction is important to provide a clear submucosal view and facilitate the dissection efficiency. To realize this, traditional ESD attaches a transparent cap on the endoscope tip (see Figure~\ref{Fig_DREAMS}(c)).

\subsection{Robot-assisted system for Endoscopic Submucosal Dissection}
To reduce the high reliance on endoscopy skills and improve the dexterity of ESD instruments, robotic technologies are applied to ESD. First, our previous work proposed a novel transendoscopic flexible parallel continuum wrist (FPCW) with three DoFs (two bending DoFs and a translational DoF) and multi-functional instruments (such as electric knives, one-DoF injection needles and two-DoF forceps). Based on the above technology, we developed a bimanual telerobotic system, called DREAMS (Dual-Arm Robotic Endoscopic Assistant for Minimally Invasive Surgery) \cite{gao2024transendoscopic, yang2024novel} to perform ESD. DREAMS consists of a patient cart and a surgeon console. The patient cart integrates the flexible endoscope, FPCW instruments and some electric devices (e.g., imaging processor, light source, electrosurgical unit, motor motion controller, etc). The surgeon console mainly comprises master devices, including master hands, pedals and buttons. During ESD procedures, the endoscopist can use the master hands to teleoperate two FPCW instruments to perform ESD bimanually. For intuitive manipulation, the endoscopist's right and left hands manipulate FPCW forceps and FPCW knife, respectively, through the direct position mapping from the end-effectors of the master hands to the FPCW instruments' tips (see Figure~\ref{Fig_DREAMS}(b)).

\section{Dataset and Benchmark}
\label{sec4}

\subsection{Multiple Motion Granularity Levels}
\label{sec4.1}

\begin{figure}
    \centering
    \includegraphics[width=5in, trim=0 10 0 0]{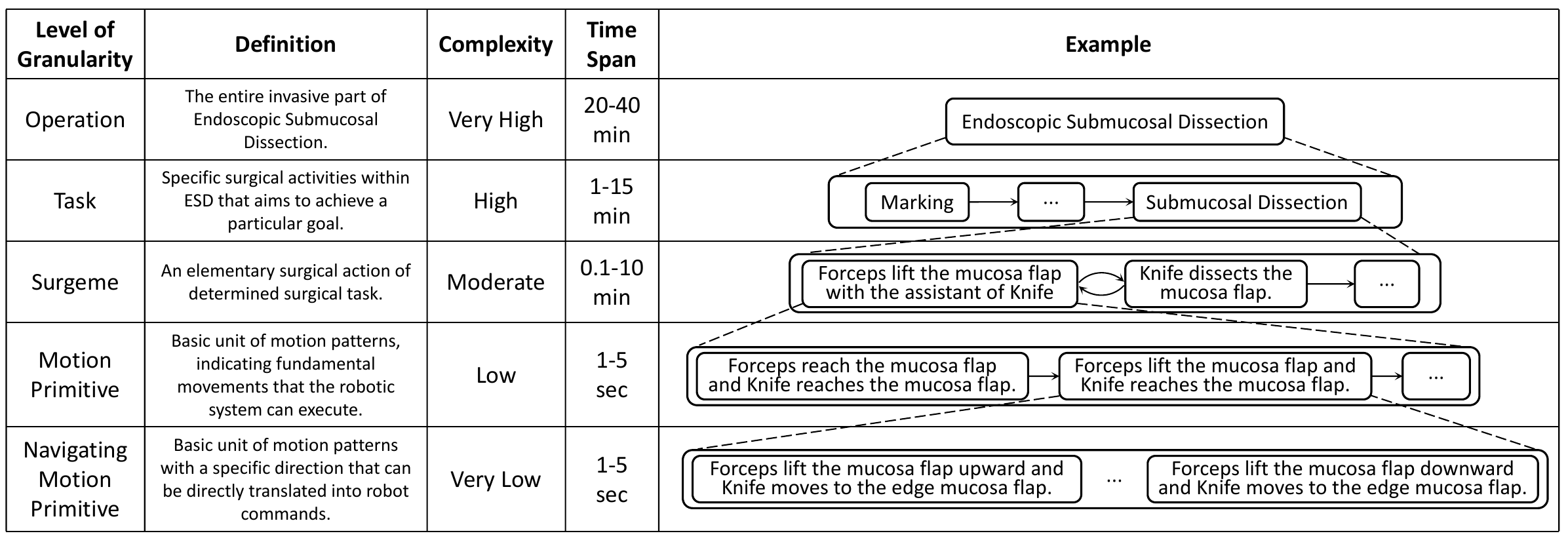}
    \caption{Overview of different levels of surgical motion granularity for Endoscopic Submucosal Dissection.}
    \label{Fig_GL}
\end{figure}

The hierarchical decomposition of surgical motion patterns is critical for advancing the ESD automation framework. Surgical interventions and the movements of the surgeon can be systematically decomposed into elements of varying granularity~\cite{vedula2016analysis,mackenzie2001hierarchical,gao2014jhu}. Although the literature has defined multiple granularity levels for certain surgical scenarios, a comprehensive and consistent framework has not yet been developed for ESD. In order to facilitate the decomposition and partial automation of ESD surgical motions, we precisely delineate and define the granularity levels based on prior research~\cite{nagy2019dvrk}, as illustrated in Figure~\ref{Fig_GL}. From a hierarchical perspective, the \textit{Operation} and \textit{Task} delineate the surgical name and workflows. The \textit{Surgeme} refers to the language instructions. Tool motions are categorized into \textit{Motion Primitive} and \textit{Navigating Motion Primitive}. Previous studies have primarily focused on task-level surgical automation, wherein tasks are executed to achieve specific goals, aligning with the concept of partial automation. These tasks can be subdivided into several surgemes, which are typically task-specific. Within the operational context of ESD, various surgemes can be constructed from a universal set of motion primitives. This notion prompts us to develop a motion primitive library comprising universal motion implementations, facilitating closer alignment with automated execution.

To construct this motion primitive library, we define the execution conditions for different surgemes. The primary execution condition is the state of the mucosal flap. For instance, if the mucosal flap has been lifted by the Forceps, the Knife will initiate the dissection (refer to the flowchart in the supplementary for additional details). Another critical condition is scene corruption during the surgeme. Given the complexity of the surgical scene, lens deviation and blurring can cause scene corruption, necessitating the cessation of all motions until the scene returns to normal. Furthermore, the direction of motion is pivotal for robot-assisted surgery. It can be predetermined based on the interaction status and relative positioning of the mucosal flaps with surgical instruments. Therefore, we introduce a navigating motion primitive to enhance the motion primitive library's alignment with automated execution, thereby providing more precise instructions to the surgeon.

\subsection{Data Construction}
\label{sec4.2}
\begin{figure}
    \centering
    \includegraphics[width=5in, trim=0 0 30 0]{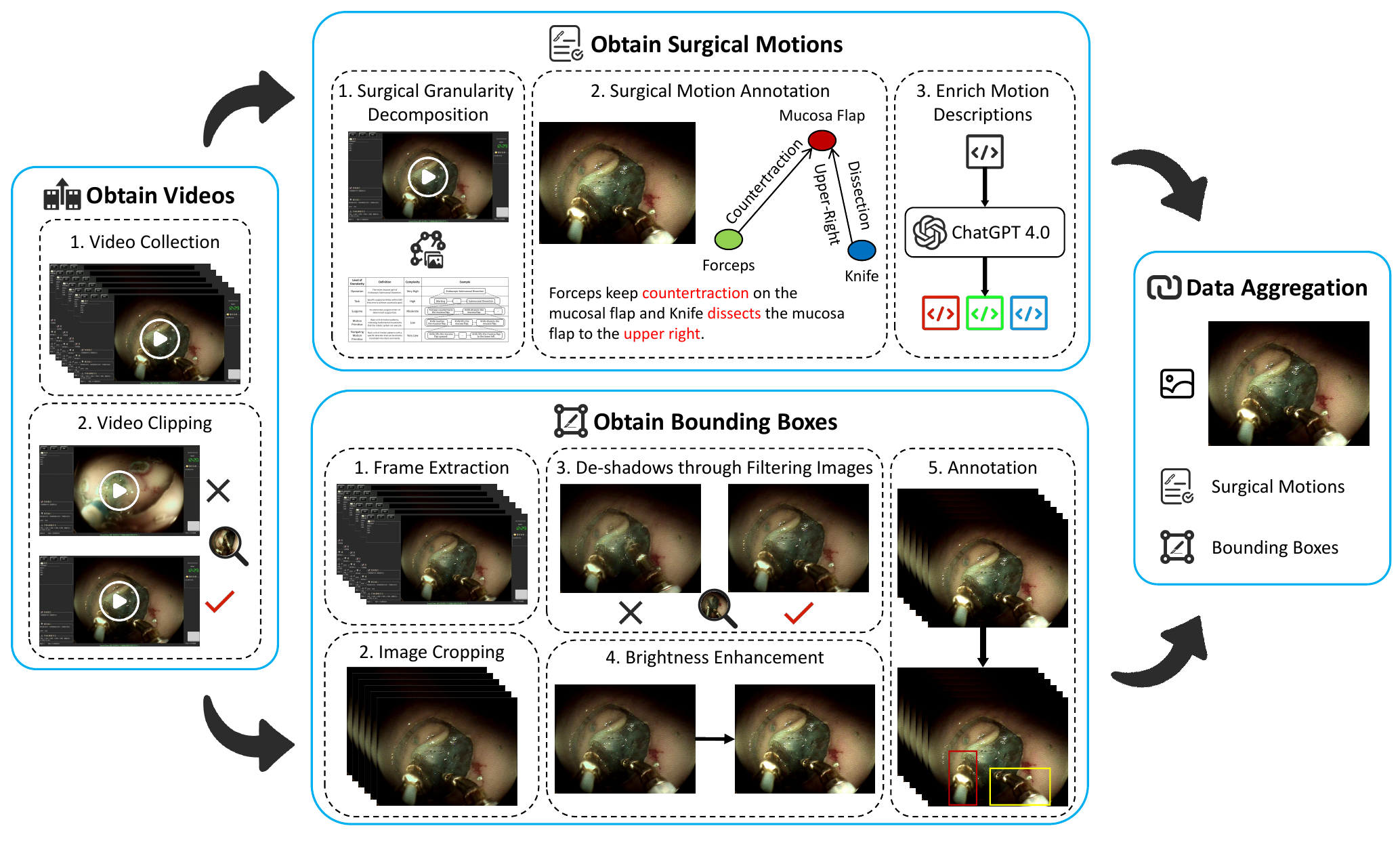}
    \caption{Overview of the construction pipeline for our CoPESD dataset, involving four key steps: video extraction, motion enrichment, bounding box annotation, and data aggregation.}
    \label{Fig_main}
\end{figure}

Our CoPESD dataset construction pipeline comprises the following steps: (1) collecting and clipping ESD videos encompassing both the DREAMS system and conventional ESD techniques, (2) extracting and enhancing images for bounding box annotation, (3) designing multi-level surgical motions and labeling each image, and (4) aggregating image and text modality data to construct the CoPESD dataset. The detailed overview of the pipeline is presented in Figure~\ref{Fig_main}.

\paragraph{Collection and Clipping of Representative ESD Videos.} The initial step of our pipeline is acquiring videos utilizing the DREAMS system and traditional ESD methods. The animal study is approved by the Institutional Ethics Committee on Animal Experiments (Approval No. DWLL-2021-021). Through these techniques, we collect 40 videos of complete robotic ESD on in-vivo porcine models, selecting 13 for the DREAMS system and 6 for conventional ESD based on video quality and surgical integrity. The ESD videos are recorded at 30 FPS with a resolution of 1920×1080. Following video preprocessing, expert endoscopists from Qilu Hospital provide the ESD surgical task annotations, encompassing six tasks: marking, injection, circumferential incision, subsidized injection, installation and debugging, and submucosal dissection. We specifically select and clip video sequences related to the submucosal dissection task, due to its high degree of soft tissue interaction and procedural complexity.
\paragraph{Image extraction and enhancement.} We sample the video sequences at 1 FPS to create the CoPESD dataset, cropping the operator interface portion to achieve a final image resolution of 1306×1009. Given the constant motion of surgical instruments within the scene, the extracted images often contain motion shadows, which could obscure key features and impair model perception. We manually eliminate images with residual shadows. Additionally, due to GI anatomical constraints and hardware limitations, ESD visual signals may suffer from insufficient illumination, complicating the learning and prediction of surgical motions. To address this, we enhance image brightness using the LLCaps~\cite{bai2023llcaps} model. Following image processing, we annotate the bounding boxes of Forceps and Knife present in each image, providing visual guidance for robot-assisted ESD.
\paragraph{Multi-level surgical motion designing and annotation.} As detailed in Section~\ref{sec4.1}, we define motion granularity levels to obtain ESD surgical motions. To ensure high-quality annotations, the following steps are conducted: first, two trained medical annotators independently annotated the images. Subsequent to the initial annotation, cross-validation is performed, and any uncertainties that arose during this process are resolved through a collaborative discussion between two experienced endoscopists. Discussions occur when the surgical scene is highly complex or the direction of dissection is not clear. After completing all annotation tasks, these two endoscopists conducted quality control of the entire dataset. Annotation evaluation depends on not only visual cues but also practical experience to decide on surgical motions and directions. Considering the inherent complexity and richness of natural language, we leverage the ChatGPT 4.0~\cite{achiam2023gpt} API to generate five distinct variants for each type of surgical motion. Each variant preserves the original semantic meaning while differing in phrasing, ensuring a diverse representation of surgical motions.

Finally, the images obtained through these steps, along with their corresponding multi-level ESD motions and bounding box annotations, are aggregated and formatted to align with the input of LVLM models~\cite{lin2023sphinx, gao2024sphinx, liu2023improved}.

\subsection{Dataset Statistics}
\begin{figure}
    \centering
    \includegraphics[width=5in, trim=0 0 30 0]{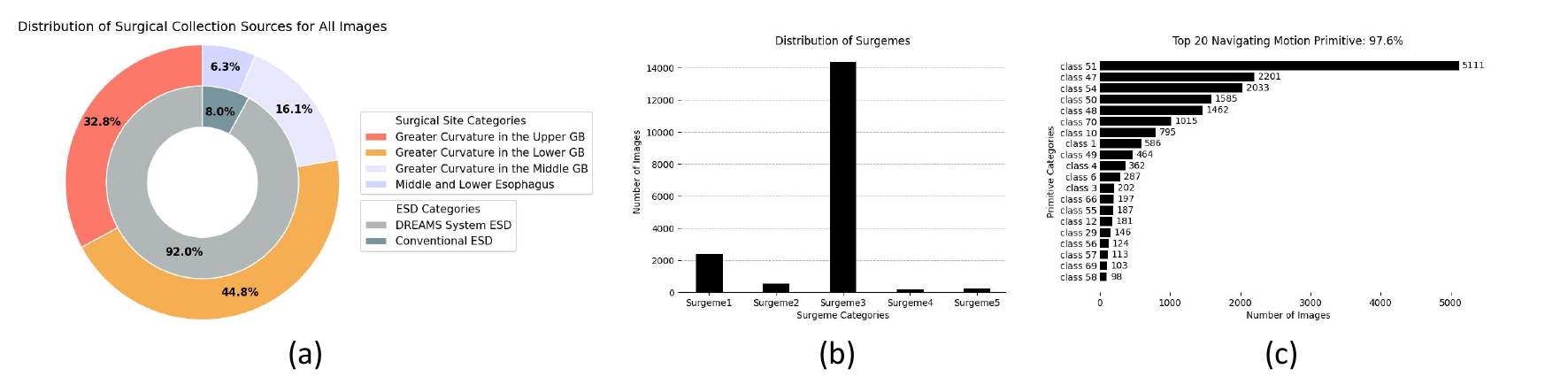}
    \caption{(a) Distribution of all images regarding the collection of surgical information. GB indicates Gastric Body. (b) Number of images within each surgeme type. (c) Distribution of entities across the top 20 navigating motion primitive types. }
    \label{Fig_stat}
\end{figure}

We contribute the CoPESD dataset, comprising 17,679 images annotated with 32,699 bounding boxes and 88,395 multi-level motions, extracted from over 35 hours of 40 ESD videos. These videos are collected from four different in-vivo porcine sites utilizing two distinct ESD techniques. The distribution overview of the extracted images is illustrated in Figure~\ref{Fig_stat} (a). The primary ESD operation sites are the Greater Curvature in the Upper and Lower Gastric Body, which constitute 77.6\% of the dataset. Due to the simpler operational steps and lower video quality of conventional ESD technique, it comprises 8\% of the dataset. Each image is annotated with multi-level granularity surgical motions specific to ESD. For the Submucosal Dissection task, we define corresponding surgemes, motion primitives, and navigating motion primitives with 5, 17, and 73 classes, respectively. Detailed descriptions for each category are provided in the supplemental material. The mean length of surgeme annotations is 5.6 words, while that for navigating motion primitive is 12.73 words (maximum = 17, minimum = 6). The distributions of these annotation types are depicted in Figures~\ref{Fig_stat} (b) and (c). Since the motion of the ESD is mainly focused on “Knife dissects the mucosa flap”, it accounts for more than 80\% of all surgemes.

\section{Experiments}
\label{sec:exp}
\subsection{Training Details}
To evaluate the effectiveness of our proposed CoPESD dataset, we conduct extensive experiments by adapting LVLMs to serve as ESD co-pilot. Specifically, we adopt state-of-the-art LVLMs, including SPHINX-X~\cite{gao2024sphinx} and LLaVA-1.5~\cite{liu2023improved}, as our baseline models. We utilize open-sourced pretrained weights and fine-tune these models on our CoPESD dataset. We refer to the fine-tuned versions of these models as SPHINX-ESD and LLaVA-ESD, respectively, in our experimental results. We experiment with two different model sizes, incorporating LLaMA-2-7B or LLaMA-2-13B~\cite{touvron2023llama} backbones. During the fine-tuning, we employ the cosine learning rate scheduler with an initial learning rate of 2e-5 and a total batch size of 64 for both SPHINX-ESD and LLaVA-ESD variants.

Moreover, we perform a thorough ablation study to assess the impact of image resolution and data proportion when fine-tuning on the proposed dataset. For the SPHINX-ESD series, we employ an input resolution of $512\times 512$, consistent with its original input image resolution for a mixed visual encoder consisting of ConvNeXt~\cite{liu2022convnet} and DINO-v2~\cite{caron2021emerging} image encoders.
We also explore an input resolution of $1024 \times 1024$, following the image partition method in SPHINX-X to break down high-resolution input images into sequences of low-resolution image patches.
For LLaVA-ESD models, we maintain the original resolution of $336\times 336$, which aligns with the setting of its visual encoder CLIP-ViT-Large~\cite{radford2021learning}. Furthermore, we investigate the effect of incorporating varying proportions of our dataset into the model fine-tuning process.
We utilize $10\%$, $50\%$, and $100\%$ of the CoPESD dataset in the overall fine-tuning to study the impact of training data on the performance of LVLMs in terms of ESD surgical instruction-following abilities.

\subsection{Evaluation Metrics}
To thoroughly assess the performance of LVLMs on the CoPESD, we select 1,787 annotations of high video quality encompassing diverse surgical tasks from our overall collections to form a comprehensive test set.
We ensure that the test set incorporates video frames from different in-vivo porcine models to guarantee a fair comparison, without any overlap with our training data. 
We design three main evaluation metrics: (1) Overall Response Quality Evaluation: We utilize GPT-4 to score the quality and accuracy of the responses generated by the fine-tuned models on a scale of 0-100 regarding surgical motions.
The prompt used to guide GPT-4 for this evaluation is provided in Appendix B.
(2) Grounding Evaluation: We assess the models' localization ability after fine-tuning on the CoPESD dataset. The mean Intersection over Union (mIoU) metric is employed to compare the ground-truth bounding boxes with the predicted ones. 
(3) Motion type and direction accuracy: We extract key items related to motion and direction from the generated responses. The evaluation metrics for this setting involve Accuracy and F-score, providing a comprehensive measure of performance. We provide the detailed definations of our evaluation metrics in Appendix C.

\begin{sidewaystable}
\caption{Quantitative comparison between different LVLM models using the CoPESD dataset.}
\label{tab1}
\centering
\renewcommand\arraystretch{1.5}
\begin{tabular}{ccccccccc}
\toprule[1pt]
\multirow{2}{*}{Model}  & \multirow{2}{*}{Image Resolution} & \multirow{2}{*}{LLM Backbone} & \multicolumn{3}{c}{GPT Score}                                  & \multicolumn{3}{c}{mIoU}                               \\ \cmidrule{4-9} 
                        &                                    &                               & \multicolumn{1}{c}{10$\%$shot} & \multicolumn{1}{c}{50$\%$shot} & \multicolumn{1}{c}{100$\%$shot} & \multicolumn{1}{c}{10$\%$shot} & \multicolumn{1}{c}{50$\%$shot} & \multicolumn{1}{c}{100$\%$shot} 
                         \\ \hline
                         
                         \multirow{2}{*}{LLaVA-ESD}  & \multirow{2}{*}{$336^{2}$}               & LLaMA2-7B                     
& 83.44 & 84.03 & 83.98 & 30.80 & 59.22 & 60.23     \\ \cmidrule{3-9} 
                        &                                    & LLaMA2-13B       
                        & 83.65 & 84.43& 83.62 & 48.00 & 61.94 & 59.42                                        \\ \hline
\multirow{4}{*}{SPHINX-ESD} & \multirow{2}{*}{$512^{2}$}               & LLaMA2-7B                     
& 84.59 & 85.03 & 84.32 & 70.08 & 69.38 & 67.38                         \\ \cmidrule{3-9} 
                        &                                    & LLaMA2-13B                   
                        & 84.53 & 85.12 & 85.39 & 69.24 & 68.63 & 69.18                             \\ \cmidrule{2-9} 
                        & \multirow{2}{*}{$1024^{2}$}              & LLaMA2-7B                     
 & 84.69 & 85.14 &  85.22 & 67.53 & 70.00 & 69.35                                            \\ \cmidrule{3-9} 
                        &                                    & LLaMA2-13B                  
        & 84.94 & 84.80 & 85.63 & 71.01 & 70.02 & 70.48                   \\ 
 \toprule[1pt]
\end{tabular}
\end{sidewaystable}

\subsection{Performance Evaluation}
\label{PE}
\paragraph{Quantitative Results} We present the quantitative performance comparisons between different LVLMs of various sizes in Table~\ref{tab1}. Utilizing our CoPESD dataset, the LVLMs exhibit substantial capabilities for surgical motion prediction across various model sizes and image resolution settings. When using the full dataset, SPHINX-ESD and LLaVA-ESD achieve the highest GPT scores of 83.98 and 85.63, respectively, demonstrating that the LVLMs can generate high-quality and accurate responses regarding surgical motions. These results suggest the effectiveness of the proposed dataset in adapting LVLMs for surgical motion generation. Furthermore, the models show significant grounding abilities, accurately localizing the positions of surgical instruments. With the full dataset, SPHINX-ESD and LLaVA-ESD achieve mIoU of 70.48 and 60.23, respectively. These results highlight the enhanced surgical instrument localization and instruction-following abilities of these models by training on the CoPESD. We notice that SPHINX-ESD outperforms LLaVA-ESD counterparts with the same LLM backbones. This may be attributed to its higher input image resolution and the more comprehensive knowledge gained through previous training.

To further assess the surgical motions generated by LVLMs, we conduct a quantitative comparison focusing on motion and direction keywords. The experimental results are presented in Table~\ref{tab2}. Fine-tuned on the full dataset, the LVLMs exhibit a robust capability in predicting surgical motions and orientations. Notably, the SPHINX-ESD model demonstrates significant improvements in accuracy and F-score for both key items with the increase in image resolution from $512\times 512$ to $1024\times 1024$. This suggests that higher image resolution enhances the model's ability to capture intricate details of surgical scenes. Additionally, utilizing a larger language model backbone generally enhanced the performance across the metrics. Overall, the trends observed in the metrics for different models regarding the keywords align with the trends of the GPT scores, indicating that the LVLMs precisely follow low-level surgical robot instructions, accurately interpreting both command structures and detailed descriptions.
\paragraph{Ablation Study} We analyze the impact of image resolution on the models' performance. As shown in Table~\ref{tab1}, increasing the input image resolution improves both the quality of generated responses and grounding accuracy. Higher resolution allows the models to capture more details in the input images, which is crucial for generating precise surgical motions. Additionally, higher resolutions align more closely with our collected video frames, preserving fine-grained information from the visual features. Besides, We explore the effects of varying amounts of sampled data for fine-tuning. Utilizing only 10$\%$ of the dataset still produces high-quality responses, comparable to those obtained with the full dataset. This demonstrates the robustness of our CoPESD dataset, as even a small subset effectively enhances the LVLMs' capability to understand and generate detailed surgical motions.

\begin{figure}
    \centering
    \includegraphics[width=5in, trim=0 0 0 0]{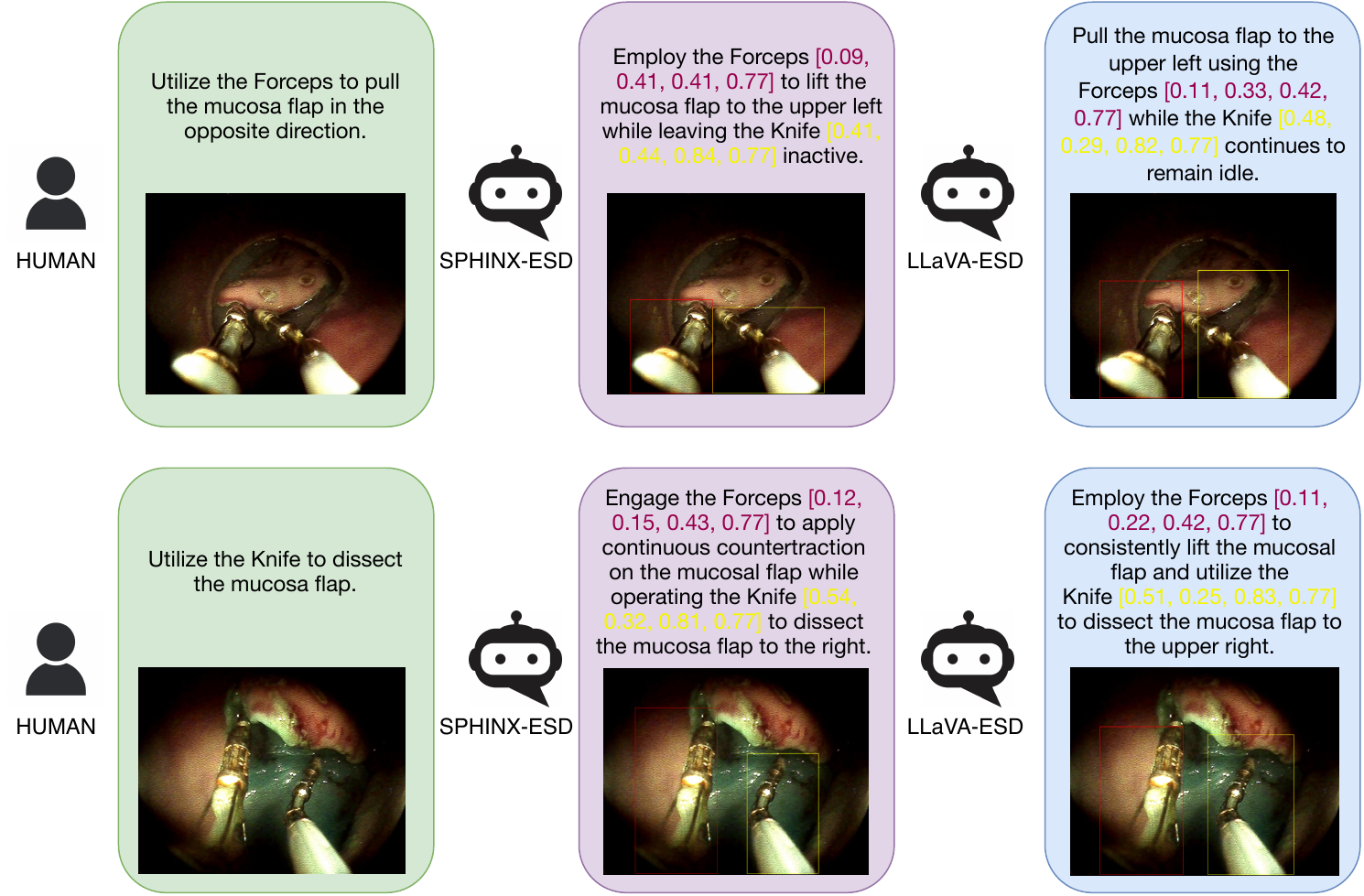}
    \caption{Demonstrations of the output surgical robot actions from LVLMs after fine-tuning on the proposed CoPESD dataset.}
    \label{fig:demo}
\end{figure}

\paragraph{Demonstrations}
We demonstrate the capabilities of LVLMs to function as an ESD co-pilot, generating intelligent surgical robot motions, as shown in Figure~\ref{fig:demo}. Both the SPHINX-ESD and LLaVA-ESD models can precisely interpret and analyze human instructions, providing clear next-step motion descriptions. Furthermore, the models accurately localize surgical instruments, specifying the exact motion and direction for each instrument, as well as detailing the comprehensive interactions between the instruments.

\section{Conclusion}
In this work, we introduce CoPESD, a comprehensive multi-level surgical motion dataset tailored for Endoscopic Submucosal Dissection. CoPESD encompasses a detailed hierarchical decomposition of ESD motions, facilitating advanced surgical automation and precision. Through rigorous quantitative analyses on state-of-the-art LVLMs, we demonstrate superior performance in predicting surgical motions provided by CoPESD, which makes LVLMs the ESD co-pilot. By making CoPESD publicly accessible, we aim to promote further research and development in the field of motion instruction-following and surgical automation, thereby advancing the integration of AI in surgical practices. 

\begin{table}[]
\caption{Quantitative full-shot performance comparison of motion type and direction generation.}
\label{tab2}
\centering
\renewcommand\arraystretch{1.3}
\begin{tabular}{ccccccc}
\toprule[1pt]
\multirow{2}{*}{Model}  & \multirow{2}{*}{Image Resolution} & \multirow{2}{*}{LLM Backbone} & \multicolumn{2}{c}{Motion}                                  & \multicolumn{2}{c}{Direction}                               \\ \cmidrule{4-7} 
                        &                                    &                               & \multicolumn{1}{c}{Accuracy} & \multicolumn{1}{c}{F-score} & \multicolumn{1}{c}{Accuracy} & \multicolumn{1}{c}{F-score} \\ \hline
\multirow{2}{*}{LLaVA-ESD}  & \multirow{2}{*}{$336^{2}$}               & LLaMA2-7B                    & 0.8716                               &  0.5143                            & 0.5833                           &      0.3429                       \\ \cmidrule{3-7} 
                        &                                    & LLaMA2-13B                    &   0.8690                            &  0.5022                            &  0.5814                             &   0.3439                         \\ \hline
\multirow{4}{*}{SPHINX-ESD} & \multirow{2}{*}{$512^{2}$}               & LLaMA2-7B                     & 0.8766   &  0.5048     &  0.5894       &    0.3561                         \\ \cmidrule{3-7} 
                        &                                    & LLaMA2-13B                    & 0.8870   & 0.6146      &  0.6024       &    0.5043                         \\ \cmidrule{2-7} 
                        & \multirow{2}{*}{$1024^{2}$}              & LLaMA2-7B                     &    0.8814                           &   0.5563                           & 0.6301                              &       0.3851                      \\ \cmidrule{3-7} 
                        &                                    & LLaMA2-13B                    &   0.8906                             &   0.6198                           & 0.6312                              &      0.5295                       \\ 
 \toprule[1pt]
\end{tabular}
\end{table}

\section{Acknowledgement}
\label{ack}
This work was supported by HK RGC, Collaborative Research Fund (CRF C4026-21GF), General Research Fund (GRF 14203323, GRF 14216022, and GRF 14211420),  NSFC/RGC Joint Research Scheme N\_CUHK420/22; Shenzhen-Hong Kong-Macau Technology Research Programme (Type C) STIC Grant 202108233000303. We would like to acknowledge the help of Mr. Junyi Wang in data annotation. 

\newpage
\begin{appendices}
\section{Datasheet}
We present a datasheet~\cite{gebru2021datasheets} for documentation and responsible usage of CoPESD.

\subsection*{Motivation}
\begin{itemize}[leftmargin=*]
\item For what purpose was the dataset created? The dataset was created to train large vision-language models as the co-pilot of Endoscopic Submucosal Dissection.

\item Who created the dataset? It was created by the authors of this paper.

\item Who funded the creation of the dataset? See the Acknowledgments in Sec~\ref{ack}

\end{itemize}

\subsection*{Composition}
\begin{itemize}[leftmargin=*]
    \item What do the instances that comprise the dataset represent? The dataset consists of images with annotated bounding boxes and multi-level motions.
    
    \item How many instances are there in total? CoPESD includes 17,679 images with 32,699 bounding boxes and 88,395 multi-level motions.
    
    \item Does the dataset contain all possible instances or is it a sample (not necessarily random) of instances from a larger set? The dataset contains all surgical motions appearing in the video of collected submucosal dissection task. 
    
    \item What data does each instance consist of? Each instance consists of an image with annotated bounding boxes and multi-level motions.
    
    \item Are relationships between individual instances made explicit? Yes, relationships between individual instances are made explicit.
    
    \item Are there recommended data splits? To thoroughly assess the performance of LVLMs on the CoPESD, we select 1,787 annotations of high video quality encompassing diverse surgical tasks from our overall collections to form a comprehensive test set.
    
    \item Are there any errors, sources of noise, or redundancies in the dataset? There are no known errors, sources of noise, or redundancies in the dataset.
    
    \item Is the dataset self-contained, or does it link to or otherwise rely on external resources (e.g., websites, tweets, other datasets)? The dataset is self-contained.
    
    \item Does the dataset contain data that might be considered confidential (e.g., data that is protected by legal privilege or by doctor–patient confidentiality, data that includes the content of individuals’ non-public communications)? No.
    
    \item Does the dataset contain data that, if viewed directly, might be offensive, insulting, threatening, or might otherwise cause anxiety? No.
    
\end{itemize}

\subsection*{Collection Process}
\begin{itemize}[leftmargin=*]
    \item How was the data associated with each instance acquired? See Sec~\ref{sec4}
    
    \item What mechanisms or procedures were used to collect the data (e.g., hardware apparatus or sensor, manual human curation, software program, software API)? See Sec~\ref{sec4}
    
    \item Who was involved in the data collection process? All the authors of this paper were involved in the data collection process.
    
    \item Over what timeframe was the data collected? The final version of the dataset was generated in May 2024.
    
\end{itemize}

\subsection*{Uses}
\begin{itemize}[leftmargin=*]
    \item Has the dataset been used for any tasks already? Yes, the CoPESD has been used for training and evaluating LVLMs as the robotic Co-Pilot of Endoscopic Submucosal Dissection in this paper.
    
    \item Is there a repository that links to any or all papers or systems that use the dataset? Yes, \href{https://github.com/gkw0010/CoPESD}{https://github.com/gkw0010/CoPESD.}
    
\end{itemize}

\subsection*{Distribution}
\begin{itemize}[leftmargin=*]
    \item Will the dataset be distributed to third parties outside of the entity (e.g., company, institution, organization) on behalf of which the dataset was created? Yes, the dataset is publicly available.
    
    \item How will the dataset be distributed (e.g., tarball on website, API, GitHub)? The dataset is distributed in GitHub.
    
    \item When will the dataset be distributed? The dataset is already available.
    
    \item Will the dataset be distributed under a copyright or other intellectual property (IP) license, and/or under applicable terms of use (ToU)? The dataset is distributed under CC BY 4.0.
        
    \item Have any third parties imposed IP-based or other restrictions on the data associated with the instances? No.
    
    \item Do any export controls or other regulatory restrictions apply to the dataset or to individual instances? No.
    
\end{itemize}

\subsection*{Maintenance}
\begin{itemize}[leftmargin=*]
    \item Who is supporting/hosting/maintaining the dataset? The authors of this paper.

    \item How can the owner/curator/manager of the dataset be contacted (e.g., email address)? Please contact Guankun Wang at \url{gkwang@link.cuhk.edu.hk}.
    
    \item Is there an erratum? No.
    
    \item Will the dataset be updated (e.g., to correct labeling errors, add new instances, delete instances)? Please check \href{https://github.com/gkw0010/CoPESD}{https://github.com/gkw0010/CoPESD} for any update.
    
\end{itemize}

\section{Author Statement}
The authors will bear all responsibility in case of violation of rights.

\section{Definition Details of Motion Granularity Levels}

\begin{figure}
    \centering
    \includegraphics[width=\textwidth, trim=0 0 0 0]{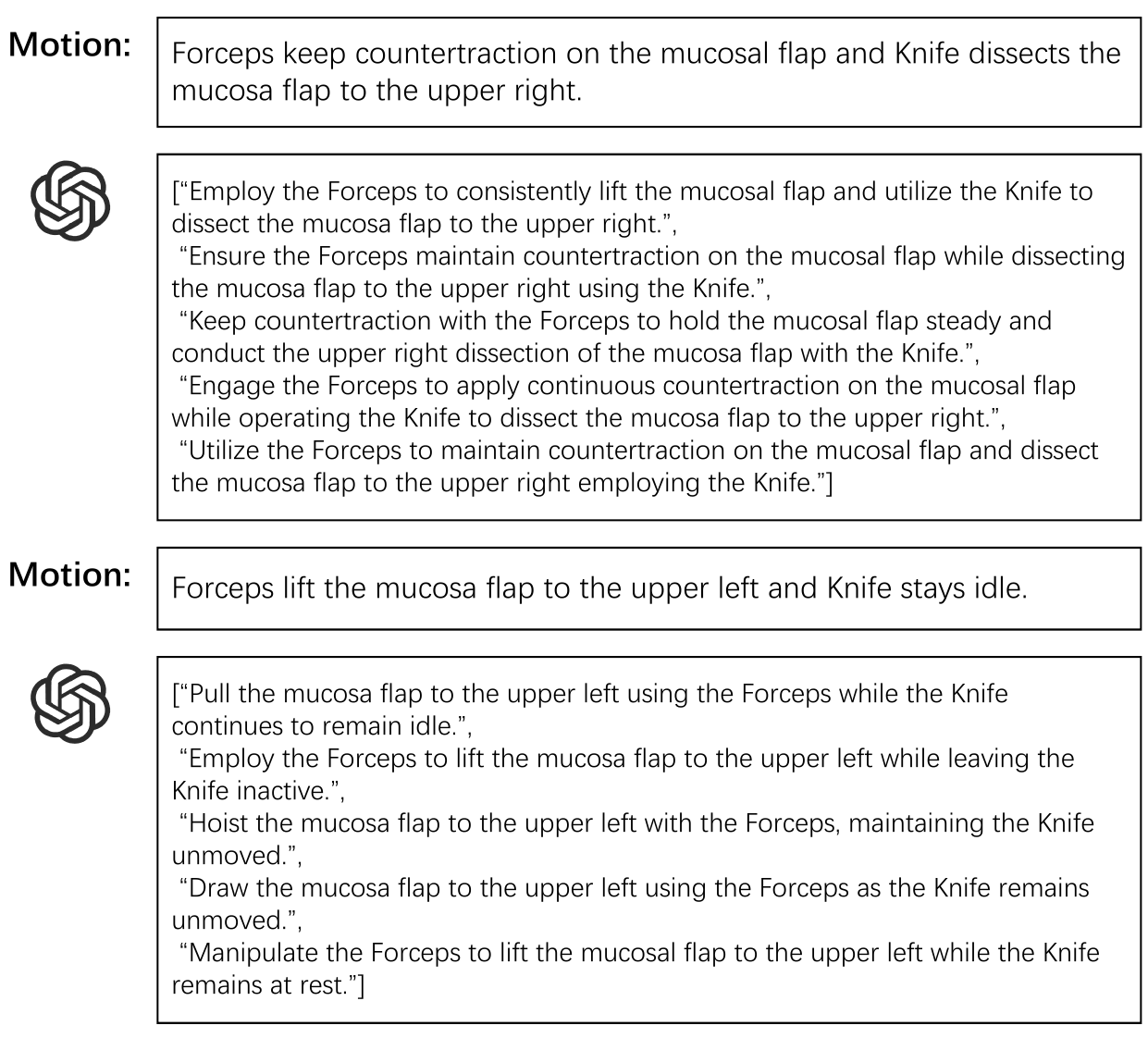}
    \caption{Ten different phrases of two “turn” motions generated by the GPT-4 API.} 
    \label{Fig_sup3}
\end{figure}

Our work divides Endoscopic Submucosal Dissection into a total of five levels of motion granularity. For applying CoPESD to LVLMs, surgeme and navigating motion primitive are utilized. Besides, we use the ChatGPT 4.0 API to generate 5 different phrases for each motion in these two levels of granularity. Figure~\ref{Fig_sup3} shows two examples. In the following, we will illustrate the definition details of the surgeme, motion primitive and navigation motion primitive.

\subsection{Definition of Surgeme Levels}
\begin{figure}
    \centering
    \includegraphics[width=\textwidth, trim=0 20 0 0]{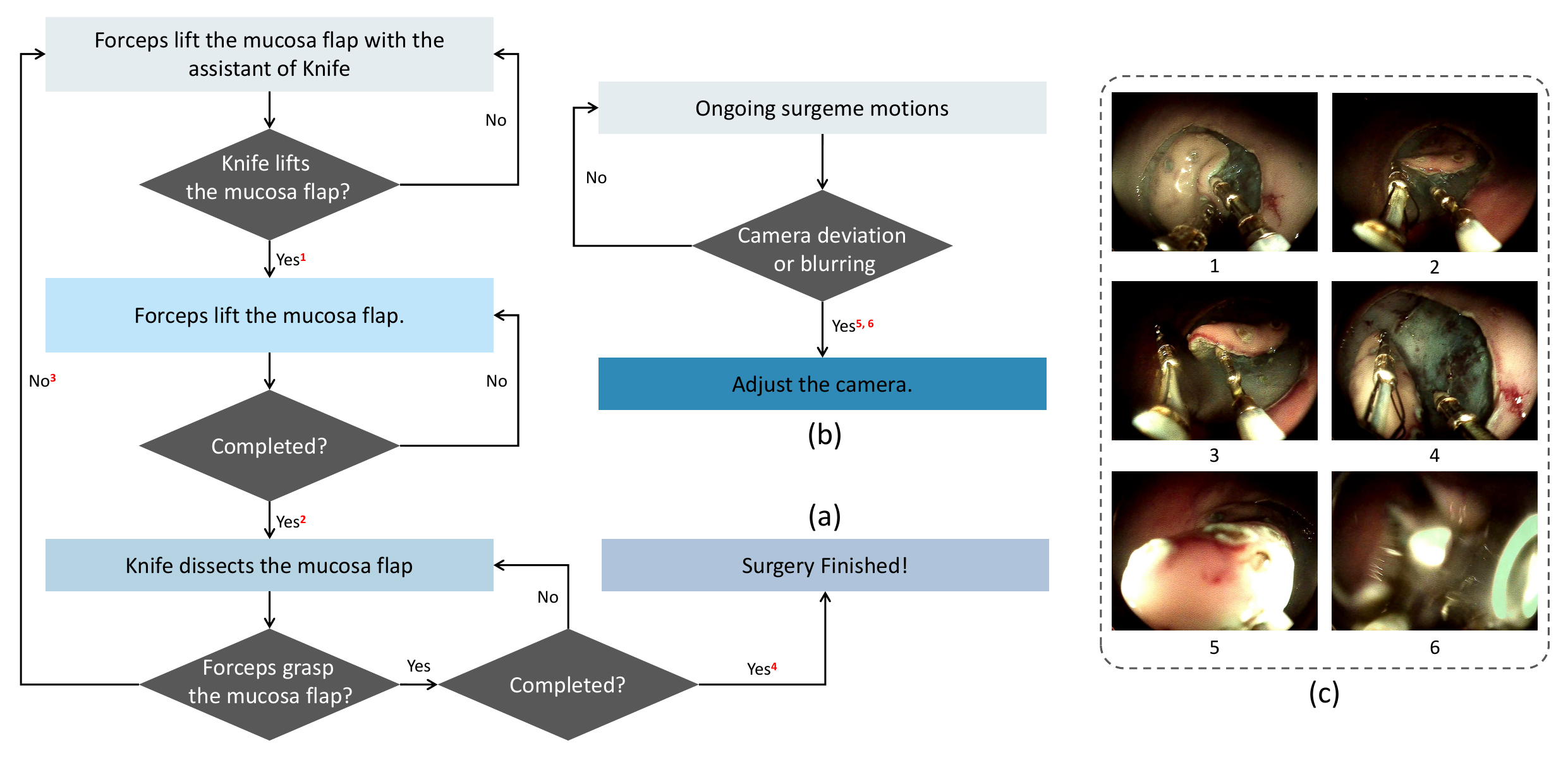}
    \caption{The flowchart and visualization of the Submucosal Dissection task. (a) Flowchart of surgeme motions. (b) Adjust the camera when deviation and blurriness occur. (c) Visualization of execution conditions.} 
    \label{Fig_sup1}
\end{figure}

Surgeme is the elementary surgical motion of determined surgical tasks. To accurately define different surgical motion granularity levels of the Submucosal Dissection task, it is essential to provide a detailed description of the procedure, progressing from simple to complex. The DREAMS system, which utilizes both Forceps and Knife as surgical instruments, relies on two instruments' coordinated operation to complete tasks. Initially, Forceps are used to lift the mucosal flap with the assistance of the Knife. The Forceps maintain the lifted state to facilitate the Knife's dissection of the mucosal flap. Throughout the procedure, the mucosal flap, covered in slippery mucus, may slip if the Forceps do not grip tightly enough. In such cases, the Knife must cease dissection and aid the Forceps in regrasping the edge of the mucosal flap. Once the Knife has elevated the mucosal flap to the optimal position, it remain idle while the Forceps independently move and grasp the edge of the mucosal flap. Upon the completion of the dissection by the Knife, the mucosal flap detaches, and the Forceps continue to grip and remove the mucosal flap. Due to the complexity of the surgical environment, deviation and blurriness in the camera view frequently occur, disrupting the surgical process. Camera adjustments are required in these instances.  Figure~\ref{Fig_sup1} illustrates this entire process and provides visualizations of each state. We summarize the aforementioned instructions and the handling of special circumstances into five surgemes, which concisely encapsulate the motions observed throughout the Submucosal Dissection task. The conventional ESD technique contains only the Knife. Therefore, the surgeme \textit{Knife dissects the mucosa flap} throughout the whole task. 

\subsection{Definition of Motion Primitive Levels}
\begin{figure}
    \centering
    \includegraphics[width=\textwidth, trim=0 20 0 0]{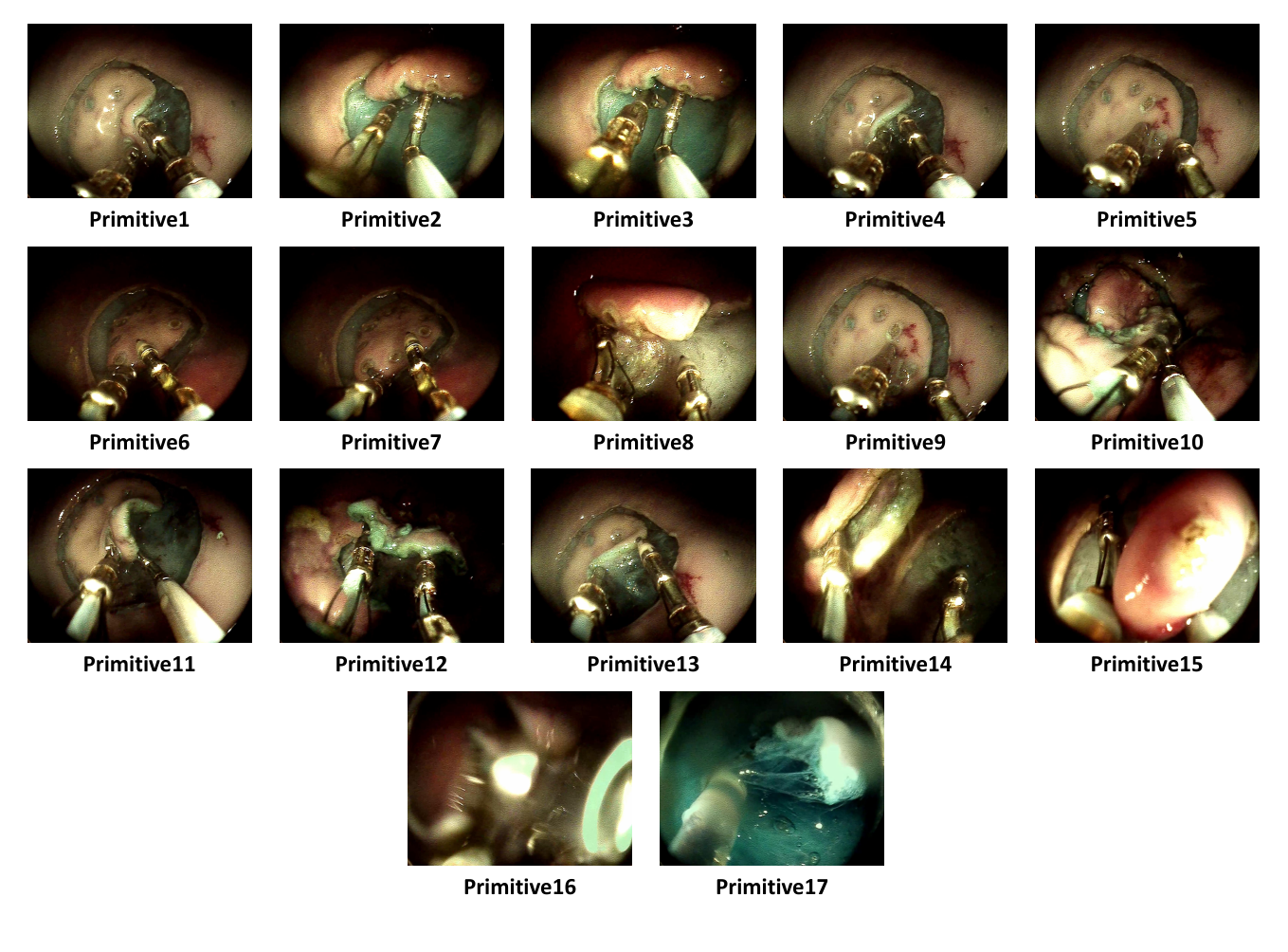}
    \caption{Visualization of 7 different types of motion primitive in CoPESD.} 
    \label{Fig_sup2}
\end{figure}

With defined surgeme, we attempt to explore the motion primitive library that makes up each surgeme from the perspective of the instrument structure. In the DREAMS system, the Knife has three degrees of freedom (DoFs): translation, pitch, and yaw. Its end-effector features a dual-knife configuration, facilitating both electrocision and electrocoagulation. Therefore, the motion primitive of the Knife contains reach (moving to the target position), lift, dissection, and idle. In contrast, the Forceps are endowed with two additional decoupled degrees of freedom, specifically allowing for rotation and grasping functionalities. Consequently, the motion primitive of the Forceps include: reach (moving to the target position), rotate, grasp, lift (or keep countertraction during Knife dissection) and idle. In the conventional ESD technique, the surgical instrument contains only the Knife. Since this instrument is similar to the structure in the DREAMS system, the motion primitive of the DREAMS system can encompass the conventional ESD technique. The full list of motion primitives and their correspondence with the surgeme can be found in Table~\ref{tab:sup2} and Table~\ref{tab:sup3}, respectively. In Figure~\ref{Fig_sup2}, we show examples of 17 different types of motion primitive.

\subsection{Definition of Navigating Motion Primitive Levels}
To enhance the alignment of the motion primitive with the robot execution language, we incorporate orientation information to create a navigating motion primitive. Based on the 2D image scene, we categorize the orientation information into eight motion directions: upward, downward, left, right, upper left, lower left, upper right, and lower right. Within the defined motion primitive, \textit{lift} and \textit{dissect} can be determined by the interaction and relative position between the surgical instruments and the mucosal flap. Conversely, \textit{reach} and \textit{rotate} are subject to randomness and angle record missing, respectively. Thus, only \textit{lift} and \textit{dissect} are assigned specific orientation information, details of which are highlighted in the Table~\ref{tab:sup4}.

\begin{table}[]
\caption{Full list of the 5 different types of surgemes in CoPESD.}
\label{tab:sup1}
\centering
\renewcommand{\arraystretch}{1.1}
\begin{tabular}{l|l}
\toprule[1pt]
\textbf{Class} & \textbf{Surgeme}                                         \\ \hline
Surgeme1       & Forceps lift the mucosa flap                           \\ \hline
Surgeme2       & Forceps lift the mucosa flap with the assistance of Knife \\ \hline
Surgeme3       & Knife dissects the mucosa flap                          \\ \hline
Surgeme4       & Forceps hold the mucosal flap                           \\ \hline
Surgeme5       & Adjust the camera                                      \\ \toprule[1pt]
\end{tabular}
\end{table}

\begin{sidewaystable}[]
\caption{Full list of the 17 different types of motion primitive in CoPESD. The bolded motions contain eight directions. Primitive17 only exists in the conventional ESD technique.}
\label{tab:sup2}
\centering
\renewcommand\arraystretch{1.5}
\begin{tabular}{l|l}
\toprule[1pt]
\textbf{Class} & \textbf{Motion Primitive}                                                                 \\ \hline
Primitive1     & Forceps reach the mucosa flap and Knife stays idle                                       \\ \hline
Primitive2     & Forceps rotate and Knife stays idle and Knife stays idle                                \\ \hline
Primitive3     & Forceps grasp the mucosa flap and Knife stays idle                                    \\ \hline
Primitive4     & \textbf{Forceps lift the mucosa flap} and Knife stays idle                               \\ \hline
Primitive5     & Forceps reach the mucosa flap and Knife reaches the mucosa flap                          \\ \hline
Primitive6     & Forceps rotate and Knife reaches the mucosa flap                                         \\ \hline
Primitive7     & Forceps grasp the mucosa flap and Knife reaches the mucosa flap                          \\ \hline
Primitive8     & \textbf{Forceps lift the mucosa flap} and Knife reaches the mucosa flap                  \\ \hline
Primitive9     & Forceps reach the mucosa flap and \textbf{Knife lifts the mucosa flap}                            \\ \hline
Primitive10    & Forceps rotate and \textbf{Knife lifts the mucosa flap}                                           \\ \hline
Primitive11    & Forceps grasp the mucosa flap and \textbf{Knife lifts the mucosa flap}                            \\ \hline
Primitive12    & Forceps stay idle and \textbf{Knife lifts the mucosa flap}                                \\ \hline
Primitive13    & Forceps keep countertraction on the mucosal flap and \textbf{Knife dissects the mucosa flap}      \\ \hline
Primitive14    & Surgical Finished! Forceps hold the mucosal flap and Knife stays idle                    \\ \hline
Primitive15    & Scenes are confusing and need to adjust the camera position! Forceps and Knife stay idle \\ \hline
Primitive16    & Scenes are confusing and need to clean the camera lens! Forceps and Knife stay idle      \\ \hline
Primitive17    & \textbf{Knife dissects the mucosa flap}                                                  \\ \toprule[1pt]
\end{tabular}
\end{sidewaystable}

\begin{sidewaystable}[]
\caption{Corresponding list between surgeme and motion primitive.}
\label{tab:sup3}
\centering
\renewcommand{\arraystretch}{1.5}
\begin{tabular}{l|l}
\toprule[1pt]
\textbf{Surgeme} & \textbf{Corresponding Motion Primitive}                                                                         \\ \hline
\multirow{4}{*}{\textbf{Surgeme1}}         & Forceps reach the mucosa flap and Knife stays idle                     \\ 
                                           & Forceps rotate and Knife stays idle and Knife stays idle                     \\ 
                                           & Forceps grasp the mucosa flap and Knife stays idle                     \\ 
                                           & Forceps lift the mucosa flap and Knife stays idle                     \\ \hline
\multirow{8}{*}{\textbf{Surgeme2}}         & Forceps reach the mucosa flap and Knife reaches the mucosa flap                          \\ 
                                           & Forceps rotate and Knife reaches the mucosa flap                                         \\ 
                                           & Forceps grasp the mucosa flap and Knife reaches the mucosa flap                          \\ 
                                           & Forceps lift the mucosa flap and Knife reaches the mucosa flap                  \\ 
                                           & Forceps reach the mucosa flap and Knife lifts the mucosa flap                            \\ 
                                           & Forceps rotate and Knife lifts the mucosa flap                                           \\ 
                                           & Forceps grasp the mucosa flap and Knife lifts the mucosa flap                            \\ 
                                           & Forceps stay idle and Knife lifts the mucosa flap             \\ \hline
\multirow{2}{*}{\textbf{Surgeme3}}         & Forceps keep countertraction on the mucosal flap and Knife dissects the mucosa flap      \\ 
                                           & Knife dissects the mucosa flap                     \\ \hline
\textbf{Surgeme4}         & Surgical Finished! Forceps hold the mucosal flap and Knife stays idle                                                                                       \\ \hline
\multirow{2}{*}{\textbf{Surgeme5}}         & Scenes are confusing and need to adjust the camera position! Forceps and Knife stay idle     \\ 
                                           & Scenes are confusing and need to clean the camera lens! Forceps and Knife stay idle                     \\  \toprule[1pt]
\end{tabular}
\end{sidewaystable}

\begin{sidewaystable}[]
\caption{List of top 20 types of navigating motion primitive in CoPESD.}
\label{tab:sup4}
\centering
\renewcommand\arraystretch{1.8}
\begin{tabular}{l|l}
\toprule[1pt]
\textbf{Class} & \textbf{Navigating Motion Primitive}                                              \\ \hline
Primitive51                & Forceps keep countertraction on the mucosal flap and Knife dissects the mucosa flap to the upper right \\ \hline
Primitive47                & Forceps keep countertraction on the mucosal flap and Knife dissects the mucosa flap upward             \\ \hline
Primitive54                & Forceps keep countertraction on the mucosal flap and Knife dissects the mucosa flap to the lower left  \\ \hline
Primitive50                & Forceps keep countertraction on the mucosal flap and Knife dissects the mucosa flap to the right       \\ \hline
Primitive48                & Forceps keep countertraction on the mucosal flap and Knife dissects the mucosa flap downward           \\ \hline
Primitive70                & Knife dissects the mucosa flap to the upper right                                                      \\ \hline
Primitive10                & Forceps lift the mucosa flap to the upper left and Knife stays idle                                    \\ \hline
Primitive1                 & Forceps move to the edge mucosa flap and Knife stays idle                                              \\ \hline
Primitive49                & Forceps keep countertraction on the mucosal flap and Knife dissects the mucosa flap to the left        \\ \hline
Primitive4                 & Forceps lift the mucosa flap upward and Knife stays idle                                               \\ \hline
Primitive6                 & Forceps lift the mucosa flap to the left and Knife stays idle                                          \\ \hline
Primitive3                 & Forceps grasp the edge of mucosa flap and Knife stays idle                                            \\ \hline
Primitive66                & Knife dissects the mucosa flap upward                                                                  \\ \hline
Primitive55                & Surgical Finished! Forceps hold the mucosal flap and Knife stays idle                                  \\ \hline
Primitive12                & Forceps move to the edge mucosa flap and Knife moves to the edge mucosa flap                           \\ \hline
Primitive29                & Forceps move to the edge mucosa flap and Knife lifts the mucosa flap to the upper left                 \\ \hline
Primitive56                & Scenes are confusing and need to adjust the camera position! Forceps and Knife stay idle               \\ \hline
Primitive57                & Scenes are confusing and need to clean the camera lens! Forceps and Knife stay idle                   \\ \hline
Primitive69                & Knife dissects the mucosa flap to the right                                                            \\ \hline
Primitive58                & Forceps stay idle and Knife lifts the mucosa flap upward                                                                    \\ \toprule[1pt]

\end{tabular}
\end{sidewaystable}

\section{Limitation and Future Work}
\label{app:limit}
The action instruction-following and motion understanding in CoPESD rely on single image inputs. This approach inherently restricts the evaluation of prediction execution during the surgical procedure, thereby undermining the reliability and efficacy of the co-pilot system. To address these limitations, future research endeavors will focus on the integration of continuous multi-frame and temporal information. By leveraging Long-Short Term Memory Networks (LSTMs) and other temporal modeling techniques, we aim to enhance the capability of LVLMs to perform more precise and context-aware surgical motion predictions. This advancement is predicated on the utilization of both historical and real-time data, facilitating a more robust and dynamic prediction framework. Such improvements are anticipated to significantly elevate the co-pilot's performance, ensuring higher reliability and better support during surgical procedures.

\section{GPT-4 Evaluation Prompts}
To evaluate the quality and accuracy of the responses generated by LLaVA-ESD and Sphinx-ESD models, we utilize the prompt below to guide GPT-4 to make comparisons between the ground-truth robotic actions and generated ones. 

\texttt{\small
{[}gt\_answer{]}\\
{[}pred\_answer{]}\\
You have been given two surgical robot instructions, the first one is the ground truth and the second one is the instruction generated by a large language model. I want you to be a surgeon who specializes in operating surgical robots. Please score (from 0 to 100) the accuracy of the instructions generated by the large language model based on the ground truth in terms of accuracy and quality. }

{[}gt\_answer{]} and {[}pred\_answer{]} represent the placeholders for ground-truth responses and predicted ones, respectively.
\section{Evaluation Metric Details for Motion and Direction}
In Section~\ref{PE}, the evaluation metrics for motion and direction involve Accuracy and F-score, providing a comprehensive measure of LVLMs' ability to capture intricate details of surgical scenes. Accuracy is a metric that measures the ratio of correctly predicted instances to the total instances. It is defined as follows:

\begin{equation}
\text{Accuracy} = \frac{TP + TN}{TP + TN + FP + FN}
\end{equation}

where: \textit{TP} (True Positives) is the number of correctly predicted positive instances. \textit{TN} (True Negatives) is the number of correctly predicted negative instances. \textit{FP} (False Positives) is the number of incorrectly predicted positive instances. \textit{FN} (False Negatives) is the number of incorrectly predicted negative instances.

Another metric, F-score (or F1-score), is the harmonic mean of Precision and Recall, providing a single measure that balances both the precision and the recall of the LVLMs. It is defined as follows:

\begin{equation}
\text{F-score} = \frac{2 \cdot \text{Precision} \cdot \text{Recall}}{\text{Precision} + \text{Recall}}
\end{equation}

where Precision and Recall are defined as:

\begin{equation}
\text{Precision} = \frac{TP}{TP + FP}
\end{equation}

\begin{equation}
\text{Recall} = \frac{TP}{TP + FN}
\end{equation}

\end{appendices}

\bibliography{sn-bibliography}


\begin{thebibliography}{58}
\ifx \bisbn   \undefined \def \bisbn  #1{ISBN #1}\fi
\ifx \binits  \undefined \def \binits#1{#1}\fi
\ifx \bauthor  \undefined \def \bauthor#1{#1}\fi
\ifx \batitle  \undefined \def \batitle#1{#1}\fi
\ifx \bjtitle  \undefined \def \bjtitle#1{#1}\fi
\ifx \bvolume  \undefined \def \bvolume#1{\textbf{#1}}\fi
\ifx \byear  \undefined \def \byear#1{#1}\fi
\ifx \bissue  \undefined \def \bissue#1{#1}\fi
\ifx \bfpage  \undefined \def \bfpage#1{#1}\fi
\ifx \blpage  \undefined \def \blpage #1{#1}\fi
\ifx \burl  \undefined \def \burl#1{\textsf{#1}}\fi
\ifx \doiurl  \undefined \def \doiurl#1{\url{https://doi.org/#1}}\fi
\ifx \betal  \undefined \def \betal{\textit{et al.}}\fi
\ifx \binstitute  \undefined \def \binstitute#1{#1}\fi
\ifx \binstitutionaled  \undefined \def \binstitutionaled#1{#1}\fi
\ifx \bctitle  \undefined \def \bctitle#1{#1}\fi
\ifx \beditor  \undefined \def \beditor#1{#1}\fi
\ifx \bpublisher  \undefined \def \bpublisher#1{#1}\fi
\ifx \bbtitle  \undefined \def \bbtitle#1{#1}\fi
\ifx \bedition  \undefined \def \bedition#1{#1}\fi
\ifx \bseriesno  \undefined \def \bseriesno#1{#1}\fi
\ifx \blocation  \undefined \def \blocation#1{#1}\fi
\ifx \bsertitle  \undefined \def \bsertitle#1{#1}\fi
\ifx \bsnm \undefined \def \bsnm#1{#1}\fi
\ifx \bsuffix \undefined \def \bsuffix#1{#1}\fi
\ifx \bparticle \undefined \def \bparticle#1{#1}\fi
\ifx \barticle \undefined \def \barticle#1{#1}\fi
\bibcommenthead
\ifx \bconfdate \undefined \def \bconfdate #1{#1}\fi
\ifx \botherref \undefined \def \botherref #1{#1}\fi
\ifx \url \undefined \def \url#1{\textsf{#1}}\fi
\ifx \bchapter \undefined \def \bchapter#1{#1}\fi
\ifx \bbook \undefined \def \bbook#1{#1}\fi
\ifx \bcomment \undefined \def \bcomment#1{#1}\fi
\ifx \oauthor \undefined \def \oauthor#1{#1}\fi
\ifx \citeauthoryear \undefined \def \citeauthoryear#1{#1}\fi
\ifx \endbibitem  \undefined \def \endbibitem {}\fi
\ifx \bconflocation  \undefined \def \bconflocation#1{#1}\fi
\ifx \arxivurl  \undefined \def \arxivurl#1{\textsf{#1}}\fi
\csname PreBibitemsHook\endcsname

\bibitem[\protect\citeauthoryear{Rudiman}{2021}]{rudiman2021minimally}
\begin{barticle}
\bauthor{\bsnm{Rudiman}, \binits{R.}}:
\batitle{Minimally invasive gastrointestinal surgery: from past to the future}.
\bjtitle{Annals of Medicine and Surgery}
\bvolume{71},
\bfpage{102922}
(\byear{2021})
\end{barticle}
\endbibitem

\bibitem[\protect\citeauthoryear{Chiu et~al.}{2021}]{chiu2021colonic}
\begin{barticle}
\bauthor{\bsnm{Chiu}, \binits{P.W.Y.}},
\bauthor{\bsnm{Ho}, \binits{K.Y.}},
\bauthor{\bsnm{Phee}, \binits{S.J.}}:
\batitle{Colonic endoscopic submucosal dissection using a novel robotic system (with video)}.
\bjtitle{Gastrointestinal Endoscopy}
\bvolume{93}(\bissue{5}),
\bfpage{1172}--\blpage{1177}
(\byear{2021})
\end{barticle}
\endbibitem

\bibitem[\protect\citeauthoryear{Bourke et~al.}{2018}]{bourke2018endoscopic}
\begin{barticle}
\bauthor{\bsnm{Bourke}, \binits{M.J.}},
\bauthor{\bsnm{Neuhaus}, \binits{H.}},
\bauthor{\bsnm{Bergman}, \binits{J.J.}}:
\batitle{Endoscopic submucosal dissection: indications and application in western endoscopy practice}.
\bjtitle{Gastroenterology}
\bvolume{154}(\bissue{7}),
\bfpage{1887}--\blpage{1900}
(\byear{2018})
\end{barticle}
\endbibitem

\bibitem[\protect\citeauthoryear{Ono et~al.}{2021}]{ono2021guidelines}
\begin{barticle}
\bauthor{\bsnm{Ono}, \binits{H.}},
\bauthor{\bsnm{Yao}, \binits{K.}},
\bauthor{\bsnm{Fujishiro}, \binits{M.}},
\bauthor{\bsnm{Oda}, \binits{I.}},
\bauthor{\bsnm{Uedo}, \binits{N.}},
\bauthor{\bsnm{Nimura}, \binits{S.}},
\bauthor{\bsnm{Yahagi}, \binits{N.}},
\bauthor{\bsnm{Iishi}, \binits{H.}},
\bauthor{\bsnm{Oka}, \binits{M.}},
\bauthor{\bsnm{Ajioka}, \binits{Y.}}, \betal:
\batitle{Guidelines for endoscopic submucosal dissection and endoscopic mucosal resection for early gastric cancer}.
\bjtitle{Digestive Endoscopy}
\bvolume{33}(\bissue{1}),
\bfpage{4}--\blpage{20}
(\byear{2021})
\end{barticle}
\endbibitem

\bibitem[\protect\citeauthoryear{Maple et~al.}{2015}]{maple2015endoscopic}
\begin{barticle}
\bauthor{\bsnm{Maple}, \binits{J.T.}},
\bauthor{\bsnm{Dayyeh}, \binits{B.K.A.}},
\bauthor{\bsnm{Chauhan}, \binits{S.S.}},
\bauthor{\bsnm{Hwang}, \binits{J.H.}},
\bauthor{\bsnm{Komanduri}, \binits{S.}},
\bauthor{\bsnm{Manfredi}, \binits{M.}},
\bauthor{\bsnm{Konda}, \binits{V.}},
\bauthor{\bsnm{Murad}, \binits{F.M.}},
\bauthor{\bsnm{Siddiqui}, \binits{U.D.}},
\bauthor{\bsnm{Banerjee}, \binits{S.}}:
\batitle{Endoscopic submucosal dissection}.
\bjtitle{Gastrointestinal Endoscopy}
\bvolume{81}(\bissue{6}),
\bfpage{1311}--\blpage{1325}
(\byear{2015})
\end{barticle}
\endbibitem

\bibitem[\protect\citeauthoryear{Cui et~al.}{2022}]{cui2022robotics}
\begin{barticle}
\bauthor{\bsnm{Cui}, \binits{Y.}},
\bauthor{\bsnm{Thompson}, \binits{C.C.}},
\bauthor{\bsnm{Chiu}, \binits{P.W.Y.}},
\bauthor{\bsnm{Gross}, \binits{S.A.}}:
\batitle{Robotics in therapeutic endoscopy (with video)}.
\bjtitle{Gastrointestinal Endoscopy}
\bvolume{96}(\bissue{3}),
\bfpage{402}--\blpage{410}
(\byear{2022})
\end{barticle}
\endbibitem

\bibitem[\protect\citeauthoryear{Odagiri and Yasunaga}{2017}]{odagiri2017complications}
\begin{botherref}
\oauthor{\bsnm{Odagiri}, \binits{H.}},
\oauthor{\bsnm{Yasunaga}, \binits{H.}}:
Complications following endoscopic submucosal dissection for gastric, esophageal, and colorectal cancer: a review of studies based on nationwide large-scale databases.
Annals of Translational Medicine
\textbf{5}(8)
(2017)
\end{botherref}
\endbibitem

\bibitem[\protect\citeauthoryear{Yamamoto et~al.}{2009}]{yamamoto2009endoscopic}
\begin{botherref}
\oauthor{\bsnm{Yamamoto}, \binits{S.}},
\oauthor{\bsnm{Uedo}, \binits{N.}},
\oauthor{\bsnm{Ishihara}, \binits{R.}},
\oauthor{\bsnm{Kajimoto}, \binits{N.}},
\oauthor{\bsnm{Ogiyama}, \binits{H.}},
\oauthor{\bsnm{Fukushima}, \binits{Y.}},
\oauthor{\bsnm{Yamamoto}, \binits{S.}},
\oauthor{\bsnm{Takeuchi}, \binits{Y.}},
\oauthor{\bsnm{Higashino}, \binits{K.}},
\oauthor{\bsnm{Iishi}, \binits{H.}}, et al.:
Endoscopic submucosal dissection for early gastric cancer performed by supervised residents: assessment of feasibility and learning curve.
Endoscopy,
923--928
(2009)
\end{botherref}
\endbibitem

\bibitem[\protect\citeauthoryear{Li et~al.}{2023}]{li2023vision}
\begin{botherref}
\oauthor{\bsnm{Li}, \binits{X.}},
\oauthor{\bsnm{Liu}, \binits{M.}},
\oauthor{\bsnm{Zhang}, \binits{H.}},
\oauthor{\bsnm{Yu}, \binits{C.}},
\oauthor{\bsnm{Xu}, \binits{J.}},
\oauthor{\bsnm{Wu}, \binits{H.}},
\oauthor{\bsnm{Cheang}, \binits{C.}},
\oauthor{\bsnm{Jing}, \binits{Y.}},
\oauthor{\bsnm{Zhang}, \binits{W.}},
\oauthor{\bsnm{Liu}, \binits{H.}}, et al.:
Vision-language foundation models as effective robot imitators.
arXiv preprint arXiv:2311.01378
(2023)
\end{botherref}
\endbibitem

\bibitem[\protect\citeauthoryear{Wang et~al.}{2024}]{wang2024surgical}
\begin{botherref}
\oauthor{\bsnm{Wang}, \binits{G.}},
\oauthor{\bsnm{Bai}, \binits{L.}},
\oauthor{\bsnm{Nah}, \binits{W.J.}},
\oauthor{\bsnm{Wang}, \binits{J.}},
\oauthor{\bsnm{Zhang}, \binits{Z.}},
\oauthor{\bsnm{Chen}, \binits{Z.}},
\oauthor{\bsnm{Wu}, \binits{J.}},
\oauthor{\bsnm{Islam}, \binits{M.}},
\oauthor{\bsnm{Liu}, \binits{H.}},
\oauthor{\bsnm{Ren}, \binits{H.}}:
Surgical-lvlm: Learning to adapt large vision-language model for grounded visual question answering in robotic surgery.
arXiv preprint arXiv:2405.10948
(2024)
\end{botherref}
\endbibitem

\bibitem[\protect\citeauthoryear{Karamcheti et~al.}{2023}]{karamcheti2023language}
\begin{botherref}
\oauthor{\bsnm{Karamcheti}, \binits{S.}},
\oauthor{\bsnm{Nair}, \binits{S.}},
\oauthor{\bsnm{Chen}, \binits{A.S.}},
\oauthor{\bsnm{Kollar}, \binits{T.}},
\oauthor{\bsnm{Finn}, \binits{C.}},
\oauthor{\bsnm{Sadigh}, \binits{D.}},
\oauthor{\bsnm{Liang}, \binits{P.}}:
Language-driven representation learning for robotics.
arXiv preprint arXiv:2302.12766
(2023)
\end{botherref}
\endbibitem

\bibitem[\protect\citeauthoryear{Fu et~al.}{2024}]{fu2024multi}
\begin{botherref}
\oauthor{\bsnm{Fu}, \binits{J.}},
\oauthor{\bsnm{Long}, \binits{Y.}},
\oauthor{\bsnm{Chen}, \binits{K.}},
\oauthor{\bsnm{Wei}, \binits{W.}},
\oauthor{\bsnm{Dou}, \binits{Q.}}:
Multi-objective cross-task learning via goal-conditioned gpt-based decision transformers for surgical robot task automation.
arXiv preprint arXiv:2405.18757
(2024)
\end{botherref}
\endbibitem

\bibitem[\protect\citeauthoryear{Furube et~al.}{2024}]{furube2024automated}
\begin{barticle}
\bauthor{\bsnm{Furube}, \binits{T.}},
\bauthor{\bsnm{Takeuchi}, \binits{M.}},
\bauthor{\bsnm{Kawakubo}, \binits{H.}},
\bauthor{\bsnm{Maeda}, \binits{Y.}},
\bauthor{\bsnm{Matsuda}, \binits{S.}},
\bauthor{\bsnm{Fukuda}, \binits{K.}},
\bauthor{\bsnm{Nakamura}, \binits{R.}},
\bauthor{\bsnm{Kato}, \binits{M.}},
\bauthor{\bsnm{Yahagi}, \binits{N.}},
\bauthor{\bsnm{Kitagawa}, \binits{Y.}}:
\batitle{Automated artificial intelligence--based phase-recognition system for esophageal endoscopic submucosal dissection (with video)}.
\bjtitle{Gastrointestinal Endoscopy}
\bvolume{99}(\bissue{5}),
\bfpage{830}--\blpage{838}
(\byear{2024})
\end{barticle}
\endbibitem

\bibitem[\protect\citeauthoryear{Huang et~al.}{2023}]{huang2023experimental}
\begin{bchapter}
\bauthor{\bsnm{Huang}, \binits{K.}},
\bauthor{\bsnm{Yuan}, \binits{X.}},
\bauthor{\bsnm{Liu}, \binits{R.}},
\bauthor{\bsnm{Zhou}, \binits{Y.}},
\bauthor{\bsnm{Hu}, \binits{B.}},
\bauthor{\bsnm{Yi}, \binits{Z.}}:
\bctitle{An experimental study of nmode in recognizing endoscopic submucosal dissection workflow}.
In: \bbtitle{2023 International Annual Conference on Complex Systems and Intelligent Science (CSIS-IAC)},
pp. \bfpage{603}--\blpage{608}
(\byear{2023}).
\bcomment{IEEE}
\end{bchapter}
\endbibitem

\bibitem[\protect\citeauthoryear{Bai et~al.}{2024}]{bai2024ossar}
\begin{botherref}
\oauthor{\bsnm{Bai}, \binits{L.}},
\oauthor{\bsnm{Wang}, \binits{G.}},
\oauthor{\bsnm{Wang}, \binits{J.}},
\oauthor{\bsnm{Yang}, \binits{X.}},
\oauthor{\bsnm{Gao}, \binits{H.}},
\oauthor{\bsnm{Liang}, \binits{X.}},
\oauthor{\bsnm{Wang}, \binits{A.}},
\oauthor{\bsnm{Islam}, \binits{M.}},
\oauthor{\bsnm{Ren}, \binits{H.}}:
Ossar: Towards open-set surgical activity recognition in robot-assisted surgery.
arXiv preprint arXiv:2402.06985
(2024)
\end{botherref}
\endbibitem

\bibitem[\protect\citeauthoryear{Cao et~al.}{2023}]{cao2023intelligent}
\begin{barticle}
\bauthor{\bsnm{Cao}, \binits{J.}},
\bauthor{\bsnm{Yip}, \binits{H.-C.}},
\bauthor{\bsnm{Chen}, \binits{Y.}},
\bauthor{\bsnm{Scheppach}, \binits{M.}},
\bauthor{\bsnm{Luo}, \binits{X.}},
\bauthor{\bsnm{Yang}, \binits{H.}},
\bauthor{\bsnm{Cheng}, \binits{M.K.}},
\bauthor{\bsnm{Long}, \binits{Y.}},
\bauthor{\bsnm{Jin}, \binits{Y.}},
\bauthor{\bsnm{Chiu}, \binits{P.W.-Y.}}, \betal:
\batitle{Intelligent surgical workflow recognition for endoscopic submucosal dissection with real-time animal study}.
\bjtitle{Nature Communications}
\bvolume{14}(\bissue{1}),
\bfpage{6676}
(\byear{2023})
\end{barticle}
\endbibitem

\bibitem[\protect\citeauthoryear{Gao et~al.}{2024}]{gao2024transendoscopic}
\begin{barticle}
\bauthor{\bsnm{Gao}, \binits{H.}},
\bauthor{\bsnm{Yang}, \binits{X.}},
\bauthor{\bsnm{Xiao}, \binits{X.}},
\bauthor{\bsnm{Zhu}, \binits{X.}},
\bauthor{\bsnm{Zhang}, \binits{T.}},
\bauthor{\bsnm{Hou}, \binits{C.}},
\bauthor{\bsnm{Liu}, \binits{H.}},
\bauthor{\bsnm{Meng}, \binits{M.Q.-H.}},
\bauthor{\bsnm{Sun}, \binits{L.}},
\bauthor{\bsnm{Zuo}, \binits{X.}}, \betal:
\batitle{Transendoscopic flexible parallel continuum robotic mechanism for bimanual endoscopic submucosal dissection}.
\bjtitle{The International Journal of Robotics Research}
\bvolume{43}(\bissue{3}),
\bfpage{281}--\blpage{304}
(\byear{2024})
\end{barticle}
\endbibitem

\bibitem[\protect\citeauthoryear{Tellex et~al.}{2020}]{tellex2020robots}
\begin{barticle}
\bauthor{\bsnm{Tellex}, \binits{S.}},
\bauthor{\bsnm{Gopalan}, \binits{N.}},
\bauthor{\bsnm{Kress-Gazit}, \binits{H.}},
\bauthor{\bsnm{Matuszek}, \binits{C.}}:
\batitle{Robots that use language}.
\bjtitle{Annual Review of Control, Robotics, and Autonomous Systems}
\bvolume{3},
\bfpage{25}--\blpage{55}
(\byear{2020})
\end{barticle}
\endbibitem

\bibitem[\protect\citeauthoryear{Kazemzadeh et~al.}{2014}]{kazemzadeh2014referitgame}
\begin{bchapter}
\bauthor{\bsnm{Kazemzadeh}, \binits{S.}},
\bauthor{\bsnm{Ordonez}, \binits{V.}},
\bauthor{\bsnm{Matten}, \binits{M.}},
\bauthor{\bsnm{Berg}, \binits{T.}}:
\bctitle{Referitgame: Referring to objects in photographs of natural scenes}.
In: \bbtitle{Proceedings of the 2014 Conference on Empirical Methods in Natural Language Processing (EMNLP)},
pp. \bfpage{787}--\blpage{798}
(\byear{2014})
\end{bchapter}
\endbibitem

\bibitem[\protect\citeauthoryear{Lu et~al.}{2019}]{lu2019vilbert}
\begin{botherref}
\oauthor{\bsnm{Lu}, \binits{J.}},
\oauthor{\bsnm{Batra}, \binits{D.}},
\oauthor{\bsnm{Parikh}, \binits{D.}},
\oauthor{\bsnm{Lee}, \binits{S.}}:
Vilbert: Pretraining task-agnostic visiolinguistic representations for vision-and-language tasks.
Advances in Neural Information Processing Systems
\textbf{32}
(2019)
\end{botherref}
\endbibitem

\bibitem[\protect\citeauthoryear{Paul et~al.}{2016}]{paul2016efficient}
\begin{botherref}
\oauthor{\bsnm{Paul}, \binits{R.}},
\oauthor{\bsnm{Arkin}, \binits{J.}},
\oauthor{\bsnm{Roy}, \binits{N.}},
\oauthor{\bsnm{M~Howard}, \binits{T.}}:
Efficient grounding of abstract spatial concepts for natural language interaction with robot manipulators
(2016)
\end{botherref}
\endbibitem

\bibitem[\protect\citeauthoryear{Shridhar and Hsu}{2018}]{shridhar2018interactive}
\begin{botherref}
\oauthor{\bsnm{Shridhar}, \binits{M.}},
\oauthor{\bsnm{Hsu}, \binits{D.}}:
Interactive visual grounding of referring expressions for human-robot interaction.
arXiv preprint arXiv:1806.03831
(2018)
\end{botherref}
\endbibitem

\bibitem[\protect\citeauthoryear{Hatori et~al.}{2018}]{hatori2018interactively}
\begin{bchapter}
\bauthor{\bsnm{Hatori}, \binits{J.}},
\bauthor{\bsnm{Kikuchi}, \binits{Y.}},
\bauthor{\bsnm{Kobayashi}, \binits{S.}},
\bauthor{\bsnm{Takahashi}, \binits{K.}},
\bauthor{\bsnm{Tsuboi}, \binits{Y.}},
\bauthor{\bsnm{Unno}, \binits{Y.}},
\bauthor{\bsnm{Ko}, \binits{W.}},
\bauthor{\bsnm{Tan}, \binits{J.}}:
\bctitle{Interactively picking real-world objects with unconstrained spoken language instructions}.
In: \bbtitle{2018 IEEE International Conference on Robotics and Automation (ICRA)},
pp. \bfpage{3774}--\blpage{3781}
(\byear{2018}).
\bcomment{IEEE}
\end{bchapter}
\endbibitem

\bibitem[\protect\citeauthoryear{Nguyen et~al.}{2020}]{nguyen2020robot}
\begin{botherref}
\oauthor{\bsnm{Nguyen}, \binits{T.}},
\oauthor{\bsnm{Gopalan}, \binits{N.}},
\oauthor{\bsnm{Patel}, \binits{R.}},
\oauthor{\bsnm{Corsaro}, \binits{M.}},
\oauthor{\bsnm{Pavlick}, \binits{E.}},
\oauthor{\bsnm{Tellex}, \binits{S.}}:
Robot object retrieval with contextual natural language queries.
arXiv preprint arXiv:2006.13253
(2020)
\end{botherref}
\endbibitem

\bibitem[\protect\citeauthoryear{Zhang et~al.}{2021}]{zhang2021invigorate}
\begin{botherref}
\oauthor{\bsnm{Zhang}, \binits{H.}},
\oauthor{\bsnm{Lu}, \binits{Y.}},
\oauthor{\bsnm{Yu}, \binits{C.}},
\oauthor{\bsnm{Hsu}, \binits{D.}},
\oauthor{\bsnm{La}, \binits{X.}},
\oauthor{\bsnm{Zheng}, \binits{N.}}:
Invigorate: Interactive visual grounding and grasping in clutter.
arXiv preprint arXiv:2108.11092
(2021)
\end{botherref}
\endbibitem

\bibitem[\protect\citeauthoryear{Mees and Burgard}{2021}]{mees2021composing}
\begin{bchapter}
\bauthor{\bsnm{Mees}, \binits{O.}},
\bauthor{\bsnm{Burgard}, \binits{W.}}:
\bctitle{Composing pick-and-place tasks by grounding language}.
In: \bbtitle{Experimental Robotics: The 17th International Symposium},
pp. \bfpage{491}--\blpage{501}
(\byear{2021}).
\bcomment{Springer}
\end{bchapter}
\endbibitem

\bibitem[\protect\citeauthoryear{Venkatesh et~al.}{2021}]{venkatesh2021spatial}
\begin{bchapter}
\bauthor{\bsnm{Venkatesh}, \binits{S.G.}},
\bauthor{\bsnm{Biswas}, \binits{A.}},
\bauthor{\bsnm{Upadrashta}, \binits{R.}},
\bauthor{\bsnm{Srinivasan}, \binits{V.}},
\bauthor{\bsnm{Talukdar}, \binits{P.}},
\bauthor{\bsnm{Amrutur}, \binits{B.}}:
\bctitle{Spatial reasoning from natural language instructions for robot manipulation}.
In: \bbtitle{2021 IEEE International Conference on Robotics and Automation (ICRA)},
pp. \bfpage{11196}--\blpage{11202}
(\byear{2021}).
\bcomment{IEEE}
\end{bchapter}
\endbibitem

\bibitem[\protect\citeauthoryear{Liu et~al.}{2022}]{liu2022structformer}
\begin{bchapter}
\bauthor{\bsnm{Liu}, \binits{W.}},
\bauthor{\bsnm{Paxton}, \binits{C.}},
\bauthor{\bsnm{Hermans}, \binits{T.}},
\bauthor{\bsnm{Fox}, \binits{D.}}:
\bctitle{Structformer: Learning spatial structure for language-guided semantic rearrangement of novel objects}.
In: \bbtitle{2022 International Conference on Robotics and Automation (ICRA)},
pp. \bfpage{6322}--\blpage{6329}
(\byear{2022}).
\bcomment{IEEE}
\end{bchapter}
\endbibitem

\bibitem[\protect\citeauthoryear{Yu et~al.}{2018}]{yu2018interactive}
\begin{botherref}
\oauthor{\bsnm{Yu}, \binits{H.}},
\oauthor{\bsnm{Zhang}, \binits{H.}},
\oauthor{\bsnm{Xu}, \binits{W.}}:
Interactive grounded language acquisition and generalization in a 2d world.
arXiv preprint arXiv:1802.01433
(2018)
\end{botherref}
\endbibitem

\bibitem[\protect\citeauthoryear{Misra et~al.}{2017}]{misra2017mapping}
\begin{botherref}
\oauthor{\bsnm{Misra}, \binits{D.}},
\oauthor{\bsnm{Langford}, \binits{J.}},
\oauthor{\bsnm{Artzi}, \binits{Y.}}:
Mapping instructions and visual observations to actions with reinforcement learning.
arXiv preprint arXiv:1704.08795
(2017)
\end{botherref}
\endbibitem

\bibitem[\protect\citeauthoryear{Anderson et~al.}{2018}]{anderson2018vision}
\begin{bchapter}
\bauthor{\bsnm{Anderson}, \binits{P.}},
\bauthor{\bsnm{Wu}, \binits{Q.}},
\bauthor{\bsnm{Teney}, \binits{D.}},
\bauthor{\bsnm{Bruce}, \binits{J.}},
\bauthor{\bsnm{Johnson}, \binits{M.}},
\bauthor{\bsnm{S{\"u}nderhauf}, \binits{N.}},
\bauthor{\bsnm{Reid}, \binits{I.}},
\bauthor{\bsnm{Gould}, \binits{S.}},
\bauthor{\bsnm{Van Den~Hengel}, \binits{A.}}:
\bctitle{Vision-and-language navigation: Interpreting visually-grounded navigation instructions in real environments}.
In: \bbtitle{Proceedings of the IEEE Conference on Computer Vision and Pattern Recognition},
pp. \bfpage{3674}--\blpage{3683}
(\byear{2018})
\end{bchapter}
\endbibitem

\bibitem[\protect\citeauthoryear{Shridhar et~al.}{2020}]{shridhar2020alfred}
\begin{bchapter}
\bauthor{\bsnm{Shridhar}, \binits{M.}},
\bauthor{\bsnm{Thomason}, \binits{J.}},
\bauthor{\bsnm{Gordon}, \binits{D.}},
\bauthor{\bsnm{Bisk}, \binits{Y.}},
\bauthor{\bsnm{Han}, \binits{W.}},
\bauthor{\bsnm{Mottaghi}, \binits{R.}},
\bauthor{\bsnm{Zettlemoyer}, \binits{L.}},
\bauthor{\bsnm{Fox}, \binits{D.}}:
\bctitle{Alfred: A benchmark for interpreting grounded instructions for everyday tasks}.
In: \bbtitle{Proceedings of the IEEE/CVF Conference on Computer Vision and Pattern Recognition},
pp. \bfpage{10740}--\blpage{10749}
(\byear{2020})
\end{bchapter}
\endbibitem

\bibitem[\protect\citeauthoryear{Shridhar et~al.}{2022}]{shridhar2022cliport}
\begin{bchapter}
\bauthor{\bsnm{Shridhar}, \binits{M.}},
\bauthor{\bsnm{Manuelli}, \binits{L.}},
\bauthor{\bsnm{Fox}, \binits{D.}}:
\bctitle{Cliport: What and where pathways for robotic manipulation}.
In: \bbtitle{Conference on Robot Learning},
pp. \bfpage{894}--\blpage{906}
(\byear{2022}).
\bcomment{PMLR}
\end{bchapter}
\endbibitem

\bibitem[\protect\citeauthoryear{Lynch and Sermanet}{2020}]{lynch2020language}
\begin{botherref}
\oauthor{\bsnm{Lynch}, \binits{C.}},
\oauthor{\bsnm{Sermanet}, \binits{P.}}:
Language conditioned imitation learning over unstructured data.
arXiv preprint arXiv:2005.07648
(2020)
\end{botherref}
\endbibitem

\bibitem[\protect\citeauthoryear{Stepputtis et~al.}{2020}]{stepputtis2020language}
\begin{barticle}
\bauthor{\bsnm{Stepputtis}, \binits{S.}},
\bauthor{\bsnm{Campbell}, \binits{J.}},
\bauthor{\bsnm{Phielipp}, \binits{M.}},
\bauthor{\bsnm{Lee}, \binits{S.}},
\bauthor{\bsnm{Baral}, \binits{C.}},
\bauthor{\bsnm{Ben~Amor}, \binits{H.}}:
\batitle{Language-conditioned imitation learning for robot manipulation tasks}.
\bjtitle{Advances in Neural Information Processing Systems}
\bvolume{33},
\bfpage{13139}--\blpage{13150}
(\byear{2020})
\end{barticle}
\endbibitem

\bibitem[\protect\citeauthoryear{Jang et~al.}{2022}]{jang2022bc}
\begin{bchapter}
\bauthor{\bsnm{Jang}, \binits{E.}},
\bauthor{\bsnm{Irpan}, \binits{A.}},
\bauthor{\bsnm{Khansari}, \binits{M.}},
\bauthor{\bsnm{Kappler}, \binits{D.}},
\bauthor{\bsnm{Ebert}, \binits{F.}},
\bauthor{\bsnm{Lynch}, \binits{C.}},
\bauthor{\bsnm{Levine}, \binits{S.}},
\bauthor{\bsnm{Finn}, \binits{C.}}:
\bctitle{Bc-z: Zero-shot task generalization with robotic imitation learning}.
In: \bbtitle{Conference on Robot Learning},
pp. \bfpage{991}--\blpage{1002}
(\byear{2022}).
\bcomment{PMLR}
\end{bchapter}
\endbibitem

\bibitem[\protect\citeauthoryear{Nair et~al.}{2022}]{nair2022learning}
\begin{bchapter}
\bauthor{\bsnm{Nair}, \binits{S.}},
\bauthor{\bsnm{Mitchell}, \binits{E.}},
\bauthor{\bsnm{Chen}, \binits{K.}},
\bauthor{\bsnm{Savarese}, \binits{S.}},
\bauthor{\bsnm{Finn}, \binits{C.}}, \betal:
\bctitle{Learning language-conditioned robot behavior from offline data and crowd-sourced annotation}.
In: \bbtitle{Conference on Robot Learning},
pp. \bfpage{1303}--\blpage{1315}
(\byear{2022}).
\bcomment{PMLR}
\end{bchapter}
\endbibitem

\bibitem[\protect\citeauthoryear{Shao et~al.}{2021}]{shao2021concept2robot}
\begin{barticle}
\bauthor{\bsnm{Shao}, \binits{L.}},
\bauthor{\bsnm{Migimatsu}, \binits{T.}},
\bauthor{\bsnm{Zhang}, \binits{Q.}},
\bauthor{\bsnm{Yang}, \binits{K.}},
\bauthor{\bsnm{Bohg}, \binits{J.}}:
\batitle{Concept2robot: Learning manipulation concepts from instructions and human demonstrations}.
\bjtitle{The International Journal of Robotics Research}
\bvolume{40}(\bissue{12-14}),
\bfpage{1419}--\blpage{1434}
(\byear{2021})
\end{barticle}
\endbibitem

\bibitem[\protect\citeauthoryear{Blukis et~al.}{2018}]{blukis2018mapping}
\begin{bchapter}
\bauthor{\bsnm{Blukis}, \binits{V.}},
\bauthor{\bsnm{Misra}, \binits{D.}},
\bauthor{\bsnm{Knepper}, \binits{R.A.}},
\bauthor{\bsnm{Artzi}, \binits{Y.}}:
\bctitle{Mapping navigation instructions to continuous control actions with position-visitation prediction}.
In: \bbtitle{Conference on Robot Learning},
pp. \bfpage{505}--\blpage{518}
(\byear{2018}).
\bcomment{PMLR}
\end{bchapter}
\endbibitem

\bibitem[\protect\citeauthoryear{Mees et~al.}{2022}]{mees2022calvin}
\begin{barticle}
\bauthor{\bsnm{Mees}, \binits{O.}},
\bauthor{\bsnm{Hermann}, \binits{L.}},
\bauthor{\bsnm{Rosete-Beas}, \binits{E.}},
\bauthor{\bsnm{Burgard}, \binits{W.}}:
\batitle{Calvin: A benchmark for language-conditioned policy learning for long-horizon robot manipulation tasks}.
\bjtitle{IEEE Robotics and Automation Letters}
\bvolume{7}(\bissue{3}),
\bfpage{7327}--\blpage{7334}
(\byear{2022})
\end{barticle}
\endbibitem

\bibitem[\protect\citeauthoryear{Nagy and Haidegger}{2019}]{nagy2019dvrk}
\begin{botherref}
\oauthor{\bsnm{Nagy}, \binits{T.D.}},
\oauthor{\bsnm{Haidegger}, \binits{T.}}:
A dvrk-based framework for surgical subtask automation.
Acta Polytechnica Hungarica,
61--78
(2019)
\end{botherref}
\endbibitem

\bibitem[\protect\citeauthoryear{Ginesi et~al.}{2021}]{ginesi2021dynamic}
\begin{barticle}
\bauthor{\bsnm{Ginesi}, \binits{M.}},
\bauthor{\bsnm{Meli}, \binits{D.}},
\bauthor{\bsnm{Roberti}, \binits{A.}},
\bauthor{\bsnm{Sansonetto}, \binits{N.}},
\bauthor{\bsnm{Fiorini}, \binits{P.}}:
\batitle{Dynamic movement primitives: Volumetric obstacle avoidance using dynamic potential functions}.
\bjtitle{Journal of Intelligent \& Robotic Systems}
\bvolume{101},
\bfpage{1}--\blpage{20}
(\byear{2021})
\end{barticle}
\endbibitem

\bibitem[\protect\citeauthoryear{Nguyen et~al.}{2019}]{nguyen2019new}
\begin{bchapter}
\bauthor{\bsnm{Nguyen}, \binits{T.}},
\bauthor{\bsnm{Nguyen}, \binits{N.D.}},
\bauthor{\bsnm{Bello}, \binits{F.}},
\bauthor{\bsnm{Nahavandi}, \binits{S.}}:
\bctitle{A new tensioning method using deep reinforcement learning for surgical pattern cutting}.
In: \bbtitle{2019 IEEE International Conference on Industrial Technology (ICIT)},
pp. \bfpage{1339}--\blpage{1344}
(\byear{2019}).
\bcomment{IEEE}
\end{bchapter}
\endbibitem

\bibitem[\protect\citeauthoryear{Wang et~al.}{2022}]{wang2022real}
\begin{barticle}
\bauthor{\bsnm{Wang}, \binits{J.}},
\bauthor{\bsnm{Jin}, \binits{Y.}},
\bauthor{\bsnm{Cai}, \binits{S.}},
\bauthor{\bsnm{Xu}, \binits{H.}},
\bauthor{\bsnm{Heng}, \binits{P.-A.}},
\bauthor{\bsnm{Qin}, \binits{J.}},
\bauthor{\bsnm{Wang}, \binits{L.}}:
\batitle{Real-time landmark detection for precise endoscopic submucosal dissection via shape-aware relation network}.
\bjtitle{Medical Image Analysis}
\bvolume{75},
\bfpage{102291}
(\byear{2022})
\end{barticle}
\endbibitem

\bibitem[\protect\citeauthoryear{Yang et~al.}{2024}]{yang2024novel}
\begin{barticle}
\bauthor{\bsnm{Yang}, \binits{X.}},
\bauthor{\bsnm{Gao}, \binits{H.}},
\bauthor{\bsnm{Fu}, \binits{S.}},
\bauthor{\bsnm{Ji}, \binits{R.}},
\bauthor{\bsnm{Hou}, \binits{C.}},
\bauthor{\bsnm{Liu}, \binits{H.}},
\bauthor{\bsnm{Luan}, \binits{N.}},
\bauthor{\bsnm{Ren}, \binits{H.}},
\bauthor{\bsnm{Sun}, \binits{L.}},
\bauthor{\bsnm{Yang}, \binits{J.}}, \betal:
\batitle{Novel miniature transendoscopic telerobotic system for endoscopic submucosal dissection (with videos)}.
\bjtitle{Gastrointestinal Endoscopy}
\bvolume{99}(\bissue{2}),
\bfpage{155}--\blpage{165}
(\byear{2024})
\end{barticle}
\endbibitem

\bibitem[\protect\citeauthoryear{Vedula et~al.}{2016}]{vedula2016analysis}
\begin{barticle}
\bauthor{\bsnm{Vedula}, \binits{S.S.}},
\bauthor{\bsnm{Malpani}, \binits{A.O.}},
\bauthor{\bsnm{Tao}, \binits{L.}},
\bauthor{\bsnm{Chen}, \binits{G.}},
\bauthor{\bsnm{Gao}, \binits{Y.}},
\bauthor{\bsnm{Poddar}, \binits{P.}},
\bauthor{\bsnm{Ahmidi}, \binits{N.}},
\bauthor{\bsnm{Paxton}, \binits{C.}},
\bauthor{\bsnm{Vidal}, \binits{R.}},
\bauthor{\bsnm{Khudanpur}, \binits{S.}}, \betal:
\batitle{Analysis of the structure of surgical activity for a suturing and knot-tying task}.
\bjtitle{PloS One}
\bvolume{11}(\bissue{3}),
\bfpage{0149174}
(\byear{2016})
\end{barticle}
\endbibitem

\bibitem[\protect\citeauthoryear{MacKenzie et~al.}{2001}]{mackenzie2001hierarchical}
\begin{barticle}
\bauthor{\bsnm{MacKenzie}, \binits{L.}},
\bauthor{\bsnm{Ibbotson}, \binits{J.}},
\bauthor{\bsnm{Cao}, \binits{C.}},
\bauthor{\bsnm{Lomax}, \binits{A.}}:
\batitle{Hierarchical decomposition of laparoscopic surgery: a human factors approach to investigating the operating room environment}.
\bjtitle{Minimally Invasive Therapy \& Allied Technologies}
\bvolume{10}(\bissue{3}),
\bfpage{121}--\blpage{127}
(\byear{2001})
\end{barticle}
\endbibitem

\bibitem[\protect\citeauthoryear{Gao et~al.}{2014}]{gao2014jhu}
\begin{bchapter}
\bauthor{\bsnm{Gao}, \binits{Y.}},
\bauthor{\bsnm{Vedula}, \binits{S.S.}},
\bauthor{\bsnm{Reiley}, \binits{C.E.}},
\bauthor{\bsnm{Ahmidi}, \binits{N.}},
\bauthor{\bsnm{Varadarajan}, \binits{B.}},
\bauthor{\bsnm{Lin}, \binits{H.C.}},
\bauthor{\bsnm{Tao}, \binits{L.}},
\bauthor{\bsnm{Zappella}, \binits{L.}},
\bauthor{\bsnm{B{\'e}jar}, \binits{B.}},
\bauthor{\bsnm{Yuh}, \binits{D.D.}}, \betal:
\bctitle{Jhu-isi gesture and skill assessment working set (jigsaws): A surgical activity dataset for human motion modeling}.
In: \bbtitle{MICCAI Workshop: M2cai},
vol. \bseriesno{3},
p. \bfpage{3}
(\byear{2014})
\end{bchapter}
\endbibitem

\bibitem[\protect\citeauthoryear{Bai et~al.}{2023}]{bai2023llcaps}
\begin{bchapter}
\bauthor{\bsnm{Bai}, \binits{L.}},
\bauthor{\bsnm{Chen}, \binits{T.}},
\bauthor{\bsnm{Wu}, \binits{Y.}},
\bauthor{\bsnm{Wang}, \binits{A.}},
\bauthor{\bsnm{Islam}, \binits{M.}},
\bauthor{\bsnm{Ren}, \binits{H.}}:
\bctitle{Llcaps: Learning to illuminate low-light capsule endoscopy with curved wavelet attention and reverse diffusion}.
In: \bbtitle{International Conference on Medical Image Computing and Computer-Assisted Intervention},
pp. \bfpage{34}--\blpage{44}
(\byear{2023}).
\bcomment{Springer}
\end{bchapter}
\endbibitem

\bibitem[\protect\citeauthoryear{Achiam et~al.}{2023}]{achiam2023gpt}
\begin{botherref}
\oauthor{\bsnm{Achiam}, \binits{J.}},
\oauthor{\bsnm{Adler}, \binits{S.}},
\oauthor{\bsnm{Agarwal}, \binits{S.}},
\oauthor{\bsnm{Ahmad}, \binits{L.}},
\oauthor{\bsnm{Akkaya}, \binits{I.}},
\oauthor{\bsnm{Aleman}, \binits{F.L.}},
\oauthor{\bsnm{Almeida}, \binits{D.}},
\oauthor{\bsnm{Altenschmidt}, \binits{J.}},
\oauthor{\bsnm{Altman}, \binits{S.}},
\oauthor{\bsnm{Anadkat}, \binits{S.}}, et al.:
Gpt-4 technical report.
arXiv preprint arXiv:2303.08774
(2023)
\end{botherref}
\endbibitem

\bibitem[\protect\citeauthoryear{Lin et~al.}{2023}]{lin2023sphinx}
\begin{botherref}
\oauthor{\bsnm{Lin}, \binits{Z.}},
\oauthor{\bsnm{Liu}, \binits{C.}},
\oauthor{\bsnm{Zhang}, \binits{R.}},
\oauthor{\bsnm{Gao}, \binits{P.}},
\oauthor{\bsnm{Qiu}, \binits{L.}},
\oauthor{\bsnm{Xiao}, \binits{H.}},
\oauthor{\bsnm{Qiu}, \binits{H.}},
\oauthor{\bsnm{Lin}, \binits{C.}},
\oauthor{\bsnm{Shao}, \binits{W.}},
\oauthor{\bsnm{Chen}, \binits{K.}}, et al.:
Sphinx: The joint mixing of weights, tasks, and visual embeddings for multi-modal large language models.
arXiv preprint arXiv:2311.07575
(2023)
\end{botherref}
\endbibitem

\bibitem[\protect\citeauthoryear{Gao et~al.}{2024}]{gao2024sphinx}
\begin{botherref}
\oauthor{\bsnm{Gao}, \binits{P.}},
\oauthor{\bsnm{Zhang}, \binits{R.}},
\oauthor{\bsnm{Liu}, \binits{C.}},
\oauthor{\bsnm{Qiu}, \binits{L.}},
\oauthor{\bsnm{Huang}, \binits{S.}},
\oauthor{\bsnm{Lin}, \binits{W.}},
\oauthor{\bsnm{Zhao}, \binits{S.}},
\oauthor{\bsnm{Geng}, \binits{S.}},
\oauthor{\bsnm{Lin}, \binits{Z.}},
\oauthor{\bsnm{Jin}, \binits{P.}}, et al.:
Sphinx-x: Scaling data and parameters for a family of multi-modal large language models.
arXiv preprint arXiv:2402.05935
(2024)
\end{botherref}
\endbibitem

\bibitem[\protect\citeauthoryear{Liu et~al.}{2023}]{liu2023improved}
\begin{botherref}
\oauthor{\bsnm{Liu}, \binits{H.}},
\oauthor{\bsnm{Li}, \binits{C.}},
\oauthor{\bsnm{Li}, \binits{Y.}},
\oauthor{\bsnm{Lee}, \binits{Y.J.}}:
Improved baselines with visual instruction tuning.
arXiv preprint arXiv:2310.03744
(2023)
\end{botherref}
\endbibitem

\bibitem[\protect\citeauthoryear{Touvron et~al.}{2023}]{touvron2023llama}
\begin{botherref}
\oauthor{\bsnm{Touvron}, \binits{H.}},
\oauthor{\bsnm{Martin}, \binits{L.}},
\oauthor{\bsnm{Stone}, \binits{K.}},
\oauthor{\bsnm{Albert}, \binits{P.}},
\oauthor{\bsnm{Almahairi}, \binits{A.}},
\oauthor{\bsnm{Babaei}, \binits{Y.}},
\oauthor{\bsnm{Bashlykov}, \binits{N.}},
\oauthor{\bsnm{Batra}, \binits{S.}},
\oauthor{\bsnm{Bhargava}, \binits{P.}},
\oauthor{\bsnm{Bhosale}, \binits{S.}}, et al.:
Llama 2: Open foundation and fine-tuned chat models.
arXiv preprint arXiv:2307.09288
(2023)
\end{botherref}
\endbibitem

\bibitem[\protect\citeauthoryear{Liu et~al.}{2022}]{liu2022convnet}
\begin{bchapter}
\bauthor{\bsnm{Liu}, \binits{Z.}},
\bauthor{\bsnm{Mao}, \binits{H.}},
\bauthor{\bsnm{Wu}, \binits{C.-Y.}},
\bauthor{\bsnm{Feichtenhofer}, \binits{C.}},
\bauthor{\bsnm{Darrell}, \binits{T.}},
\bauthor{\bsnm{Xie}, \binits{S.}}:
\bctitle{A convnet for the 2020s}.
In: \bbtitle{Proceedings of the IEEE/CVF Conference on Computer Vision and Pattern Recognition},
pp. \bfpage{11976}--\blpage{11986}
(\byear{2022})
\end{bchapter}
\endbibitem

\bibitem[\protect\citeauthoryear{Caron et~al.}{2021}]{caron2021emerging}
\begin{bchapter}
\bauthor{\bsnm{Caron}, \binits{M.}},
\bauthor{\bsnm{Touvron}, \binits{H.}},
\bauthor{\bsnm{Misra}, \binits{I.}},
\bauthor{\bsnm{J{\'e}gou}, \binits{H.}},
\bauthor{\bsnm{Mairal}, \binits{J.}},
\bauthor{\bsnm{Bojanowski}, \binits{P.}},
\bauthor{\bsnm{Joulin}, \binits{A.}}:
\bctitle{Emerging properties in self-supervised vision transformers}.
In: \bbtitle{Proceedings of the IEEE/CVF International Conference on Computer Vision},
pp. \bfpage{9650}--\blpage{9660}
(\byear{2021})
\end{bchapter}
\endbibitem

\bibitem[\protect\citeauthoryear{Radford et~al.}{2021}]{radford2021learning}
\begin{bchapter}
\bauthor{\bsnm{Radford}, \binits{A.}},
\bauthor{\bsnm{Kim}, \binits{J.W.}},
\bauthor{\bsnm{Hallacy}, \binits{C.}},
\bauthor{\bsnm{Ramesh}, \binits{A.}},
\bauthor{\bsnm{Goh}, \binits{G.}},
\bauthor{\bsnm{Agarwal}, \binits{S.}},
\bauthor{\bsnm{Sastry}, \binits{G.}},
\bauthor{\bsnm{Askell}, \binits{A.}},
\bauthor{\bsnm{Mishkin}, \binits{P.}},
\bauthor{\bsnm{Clark}, \binits{J.}}, \betal:
\bctitle{Learning transferable visual models from natural language supervision}.
In: \bbtitle{International Conference on Machine Learning},
pp. \bfpage{8748}--\blpage{8763}
(\byear{2021}).
\bcomment{PMLR}
\end{bchapter}
\endbibitem

\bibitem[\protect\citeauthoryear{Gebru et~al.}{2021}]{gebru2021datasheets}
\begin{barticle}
\bauthor{\bsnm{Gebru}, \binits{T.}},
\bauthor{\bsnm{Morgenstern}, \binits{J.}},
\bauthor{\bsnm{Vecchione}, \binits{B.}},
\bauthor{\bsnm{Vaughan}, \binits{J.W.}},
\bauthor{\bsnm{Wallach}, \binits{H.}},
\bauthor{\bsnm{Iii}, \binits{H.D.}},
\bauthor{\bsnm{Crawford}, \binits{K.}}:
\batitle{Datasheets for datasets}.
\bjtitle{Communications of the ACM}
\bvolume{64}(\bissue{12}),
\bfpage{86}--\blpage{92}
(\byear{2021})
\end{barticle}
\endbibitem

\end{thebibliography}

\end{document}